%% file: main.tex
\pdfoutput=1

\documentclass[11pt]{article}

\usepackage[final]{acl}

\usepackage{times}
\usepackage{latexsym}

\usepackage[T1]{fontenc}

\usepackage[utf8]{inputenc}

\usepackage{microtype}

\usepackage{inconsolata}

\usepackage{graphicx}

\usepackage{placeins}
\usepackage{multirow}
\usepackage{arydshln}
\usepackage{graphicx}
\usepackage{paralist}
\usepackage{amssymb}
\usepackage{amsmath}
\usepackage{pifont}
\usepackage{xcolor}
\usepackage{bbm}
\usepackage{listings}
\usepackage{xspace}
\usepackage[capitalize,noabbrev]{cleveref}
\usepackage{bbold}

\title{\benchmarkname: On-Demand Keyphrase Generation}

\author{
Di Wu\thanks{Equal contribution}, Xiaoxian Shen\footnotemark[1], Kai-Wei Chang \\
University of California, Los Angeles \\ 
\texttt{\{diwu,kwchang\}@cs.ucla.edu, xhshen@ucla.edu}
}

\newcommand{\benchmarkname}{\textsc{MetaKP}\xspace}

\begin{document}
\maketitle

\input{text/0_abstract}
\input{text/1_introduction}

\input{text/2_related_work}

\input{text/3_benchmark}

\input{text/4_approach}
\input{text/5_results}

\input{text/6_discussion}
\input{text/7_conclusion}

\bibliography{anthology, custom}


\input{text/appendix}

\end{document}

%% file: text/0_abstract.tex
\begin{abstract}

Traditional keyphrase prediction methods predict a single set of keyphrases per document, failing to cater to the diverse needs of users and downstream applications. To bridge the gap, we introduce on-demand keyphrase generation, a novel paradigm that requires keyphrases that conform to specific high-level goals or intents. For this task, we present \benchmarkname, a large-scale benchmark comprising four datasets, 7500 documents, and 3760 goals across news and biomedical domains with human-annotated keyphrases. Leveraging \benchmarkname, we design both supervised and unsupervised methods, including a multi-task fine-tuning approach and a self-consistency prompting method with large language models. The results highlight the challenges of supervised fine-tuning, whose performance is not robust to distribution shifts. By contrast, the proposed self-consistency prompting approach greatly improves the performance of large language models, enabling GPT-4o to achieve 0.548 SemF1, surpassing the performance of a fully fine-tuned BART-base model. Finally, we demonstrate the potential of our method to serve as a general NLP infrastructure, exemplified by its application in epidemic event detection from social media.

\end{abstract}

%% file: text/1_introduction.tex
\section{Introduction}

Keyphrase prediction is an NLP task that has attracted long-lasting research interest 
\citep{witten1999kea, hulth-2003-improved, meng-etal-2017-deep}. Given documents from various domains such as academic writing, news, social media, or meetings, keyphrase extraction and keyphrase generation models output short phrases aiming at encapsulating the key entities and concepts mentioned by the document. Beyond a number of information retrieval applications \citep{kim-etal-2013-applying, Tang2017QALinkET, boudin-etal-2020-keyphrase}, keyphrase prediction methods are widely incorporated into the pipelines of other NLP tasks such as natural language generation \citep{yao2019plan, DBLP:conf/nips/LiLMJLK20}, text summarization \citep{dou-etal-2021-gsum}, and text classification \citep{berend-2011-opinion}.

\begin{figure}[t!]
\includegraphics[width=\linewidth]{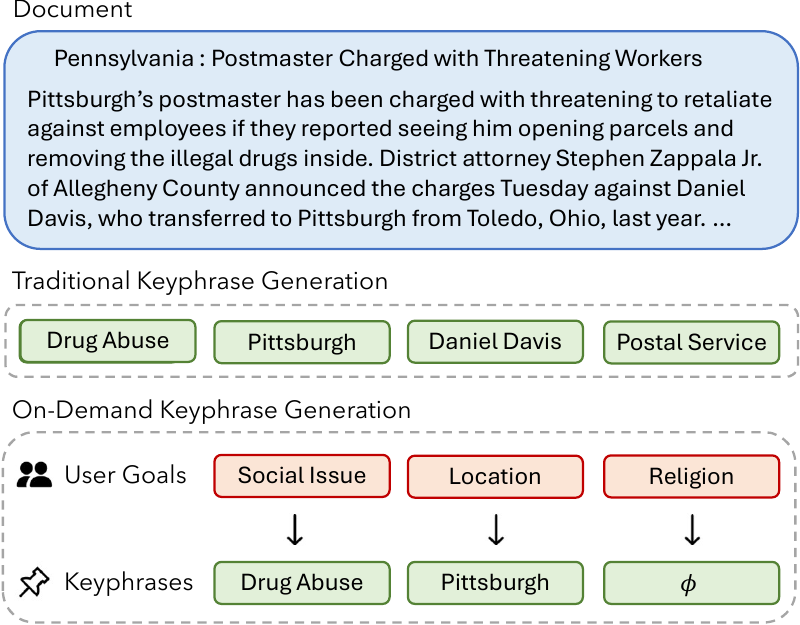}
\caption{An illustration of on-demand keyphrase generation. Given diverse user goals, models are required to generate goal-conforming keyphrases or abstain. }
\label{main-framework}
\end{figure}

Despite their wide application in diverse scenarios, which may have diverse requirements on the types of keyphrases, existing keyphrase prediction methods generally follow a suboptimal assumption: for every document, the model shall predict a \textit{single} application-agnostic set of keyphrases, which is then evaluated against a \textit{monolithic} set of references \citep{wu2023kpeval}. This one-size-fits-all approach fails to cater to both downstream applications' varied requirements of the keyphrase predictions' topic and level of specificity and different expectations from human users with diverse backgrounds. To properly handle such diverse feedback, current approach could only rely on the sample-rerank strategy \citep{zhao-etal-2022-keyphrase,wu-etal-2023-rethinking-model}, which is largely inefficient. Besides, the single-reference setting also biases the intrinsic evaluation, of keyphrase prediction models, as high-frequency topics in keyphrase labels may significantly outweigh the long-tail keyphrases.

To tackle these challenges, we propose \textit{on-demand keyphrase generation}, a novel paradigm that predicts keyphrases conditioned on a \textit{goal} phrase that specifies the high-level category or intent of the keyphrase (\cref{main-framework}). For existing keyphrase prediction models, this task is challenging as it requires the predictions to be not only capturing key information but also goal-conforming. Furthermore, the models are required to accept \textit{open-vocabulary} goals, a significant step beyond predicting keyphrases with predefined categories or ontology \citep{park-caragea-2023-multi}.

To test on this new task, we meticulously curate and release \benchmarkname, a large-scale on-demand keyphrase generation benchmark covering four datasets, 7500 documents, and 3760 unique goals from the news and the biomedical text domain. We build a scalable labeling pipeline that combines GPT-4 \citep{OpenAI2023GPT4TR} and human annotators to construct high-quality goals from keyphrases (\cref{benchmark-construction}). For evaluation, we design two tasks: judging the relevance of goals and generating goal-conforming keyphrases. For the latter, we employ the state-of-the-art evaluation method \citep{wu2023kpeval} to conduct a semantic-based evaluation. 

Using \benchmarkname, we develop both supervised and unsupervised methods for on-demand keyphrase generation. For the supervised method, we design a multi-task fine-tuning approach to enable sequence-to-sequence pre-trained language models to self-determine the relevance of a goal and selectively generate keyphrases (\cref{section-approach-supervised}). Then, in \cref{section-approach-unsupervised}, we introduce an unsupervised self-consistency prompting approach leveraging the strong ability of large language models (LLMs) to propose goal-related keyphrase candidates and their propensity to predict high quality keyphrases with higher frequencies and ranks. Comprehensive experiments reveal the following insights:

\begin{compactenum}
    \item \benchmarkname represents a challenging benchmark for keyphrase generation. Flan-T5-XL, the strongest fine-tuned model, only achieves an average of 0.609 Satisfaction Rate across all the datasets, and zero-shot prompting GPT-4o, a strong LLM, only achieves 0.492 SR. 
    \item The proposed fine-tuning approach enables jointly learning goal relevance judgment and keyphrase generation without impeding each task's performance (\cref{section-analyses}). 
    \item The proposed self-consistency prompting approach greatly improves the performance of LLMs, enabling GPT-4o to achieve 0.548 SemF1, surpassing the performance of a fully fine-tuned BART-base model. 
    \item Supervised fine-tuning can fail to generalize on out-of-distribution testing data. By contrast, LLM-based unsupervised method achieves consistent performance in all the domains, especially in the news domain, where GPT-4o outperforms supervised Flan-T5-XL by 19\% in out-of-distribution testing. 
    
\end{compactenum}

Finally, we demonstrate the potential of on-demand keyphrase generation as a general NLP infrastructure. Specifically, we use event detection for epidemics prediction \citep{parekh2024event} as a test bed. By constructing simple goals from event ontology and attempting to extract relevant keyphrases from social media text, we show that an on-demand keyphrase generation model has the potential to extract epidemic-related trends similar to an event detection model trained on task-specific data. The benchmark and experimental code will be released at \url{https://github.com/uclanlp/MetaKP} to facilitate further research.

%% file: text/2_related_work.tex
\begin{figure*}[ht!]
\centering
\includegraphics[width=0.95\linewidth]{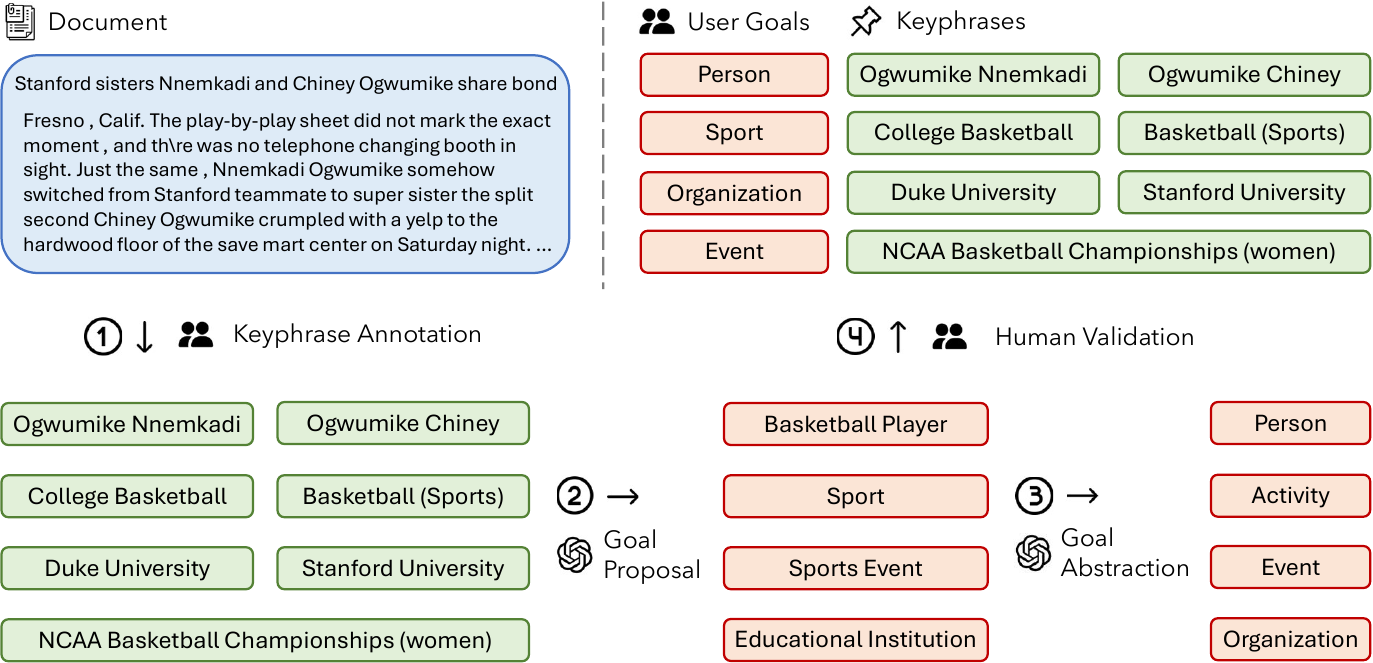}
\caption{The annotation pipeline for \benchmarkname. Starting from human-annotated keyphrases, GPT-4 is instructed to propose high-level goals and self-refine them. Finally, the goals are validated and filtered by humans.}
\label{benchmark-construction}
\end{figure*}

\section{Related Work}

\paragraph{Keyphrase Prediction with Types} This work is closely related to prior work on modeling keyphrases with pre-defined types or categories. Early datasets are often derived from named entity recognition, where keyphrase spans are extracted with entity type tags \citep{qasemizadeh-schumann-2016-acl, augenstein-etal-2017-semeval,luan-etal-2018-multi}. Notable modeling approaches include using intermediate task for training strong and transferable encoder representations \citep{park-caragea-2020-scientific} as well as multi-task fine-tuning \citep{park-caragea-2023-multi}. In addition, existing literature has explored inducing high-level type variable for more accurate keyphrase prediction, such as topic-guided keyphrase generation \citep{wang-etal-2019-topic-aware,DBLP:conf/sigir/ZhangJY0W22}, hierarchical keyphrase generation \citep{wang-etal-2016-extracting, chen-etal-2020-exclusive, DBLP:conf/pkdd/ZhangYJLW22}, as well as keyphrase completion \citep{deepkpcompletion}. Compare to these prior work, our benchmark features a massive set of open-vocabulary goals with wide domain coverage. We design novel supervised and unsupervised modeling approaches that consider up-to-date techniques such as large language models.

\paragraph{On-Demand Information Extraction} Our work resonates with the recent trend of designing flexible formulations for information extraction. For instance, \citet{zhong-etal-2021-qmsum} propose a query-focused formulation for the summarization task, and \citet{zhang-etal-2023-macsum} further extend the task to include five constraints: Length, Extractiveness, Specificity, Topic, and Speaker. Recently, \citet{jiao-etal-2023-instruct} introduce on-demand information extraction, where models are required to answer queries by extracting information from the associated text and organize it in a tabular format. By comparison, this work pioneers in defining and benchmarking the goal-following ability of keyphrase prediction models. Our resource and methodology lay the foundation for user-controllable keyphrase systems and flexible concept extraction infrastructures.

%% file: text/3_benchmark.tex
\section{\benchmarkname Benchmark}
\label{section-benchmark}

In this section, we formulate the on-demand keyphrase generation task and introduce the \benchmarkname evaluation benchmark.

\subsection{Problem Formulation}
The traditional keyphrase prediction task is defined with a tuple: (document $\mathcal{X}$, reference set $\mathcal{Y}$). Given $\mathcal{X}$, a model directly generates all keyphrase hypotheses, with approximating $\mathcal{Y}$ as the goal. For on-demand keyphrase generation, we introduce an open-vocabulary goal phrase $g$ which describes a category of keyphrases specified by the user. The target of the model, then, is to generate a set of keyphrases based on $(\mathcal{X}, g)$ to approximate the set of goal-conforming keyphrases $\mathcal{Y}_g\subseteq\mathcal{Y}$. 

\cref{main-framework} provides an intuitive example of the task. We note that for irrelevant goals,  $\mathcal{Y}_g = \phi$, which means that an ideal model should not generate any keyphrases given such goals. In addition, although $\mathcal{Y}_g$ varies according to the goal, the universal set of keyphrases $\mathcal{Y}$ is assumed to be generally fixed. In other words, $g$ could be viewed as a query that specifies a target subset from $\mathcal{Y}$, which enables a wide range of choices for the modeling design. 

\subsection{Benchmark Creation Pipeline}

To evaluate on-demand keyphrase generation, we curate \benchmarkname, a large-scale multi-domain evaluation benchmark. The key challenge is to construct general, meaningful, and diverse goals that reflect high-level keyphrase types in real-world scenarios such as document indexing and search engines. To collect high quality goals, we design a model-in-the-loop annotation pipeline that combines GPT-4 \citep{OpenAI2023GPT4TR} with human annotators to infer goals reversely from keyphrase annotations (\cref{benchmark-construction}), with four steps detailed as follows.

\paragraph{Keyphrase Annotation by Human} Given the document $\mathcal{X}$, human annotators specify the set of all the possible keyphrases $\mathcal{Y}$. For \benchmarkname, we directly leverage the expert-curated keyphrases from the respective keyphrase prediction datasets.

\paragraph{Goal Proposal} We instruct GPT-4 to propose a high-level goal for each of the keyphrases, and the same goal could be shared by multiple keyphrases\footnote{We use \texttt{gpt-4-0613} via the OpenAI API.}. Concretely, given $\mathcal{X}, \mathcal{Y}$, GPT-4 returns a mapping from goals to keyphrases. We present the prompt for this step in \cref{appendix-dataset-details}. 

\paragraph{Goal Abstraction} After the previous step, a draft goal has been associated with each keyphrase. Although the proposed goals are relevant, we observe that they are sometimes overly specific. Therefore, we instruct GPT-4 to perform a round of \textit{self-refinement}, where it attempts to propose a more abstract version for each of the goals in the previous round, or keep the original goals if they are already high-level enough. The full prompt for this step is presented in \cref{appendix-dataset-details}. 

\paragraph{Human Validation} We qualitatively find that the outputs from two GPT-4 annotation iterations are sufficiently abstract and diverse. To further improve the quality of the goals and reduce the level of duplication, two of the authors conduct a round of filtering to obtain the final goal annotations. As this step does not entail adding new goals, the annotators achieve a high inter-annotator agreement (detailed in the next section) following the annotation guideline, which we present in \cref{appendix-dataset-details}. Finally, we create an instance for each of the filtered goals, taking the form $(\mathcal{X}, g_i, \mathcal{Y}_{g_i})$.

\subsection{Dataset Statistics}

We execute the aforementioned goal construction pipeline on four keyphrase prediction datasets covering two domains: news and biomedical text. For each domain, we create both an in-distribution and an out-of-distribution test set.

\begin{itemize}
    \item \textbf{KPTimes} \citep{gallina-etal-2019-kptimes} is a large-scale keyphrase generation dataset in the news domain. The documents are sourced from from New York Times and the keyphrases are curated by professional editors.
    \item \textbf{DUC2001} \citep{wan2008single} is a widely used keypharse extraction dataset with news articles collected from TREC-9, paired with human-annotated keyphrases.
    \item \textbf{KPBiomed} \citep{houbre-etal-2022-large} is a large-scale dataset containing PubMed abstracts paired with keyphrases annotated by paper authors themselves.
    \item \textbf{Pubmed} \citep{Schutz2008KeyphraseEF} is a traditional keyphrase extraction dataset in the biomedical domain with documents and keyphrases extracted from the PubMed Central. 
\end{itemize}

\input{tables/dataset_stats}

\begin{figure}[t!]
\centering
\includegraphics[width=\linewidth]{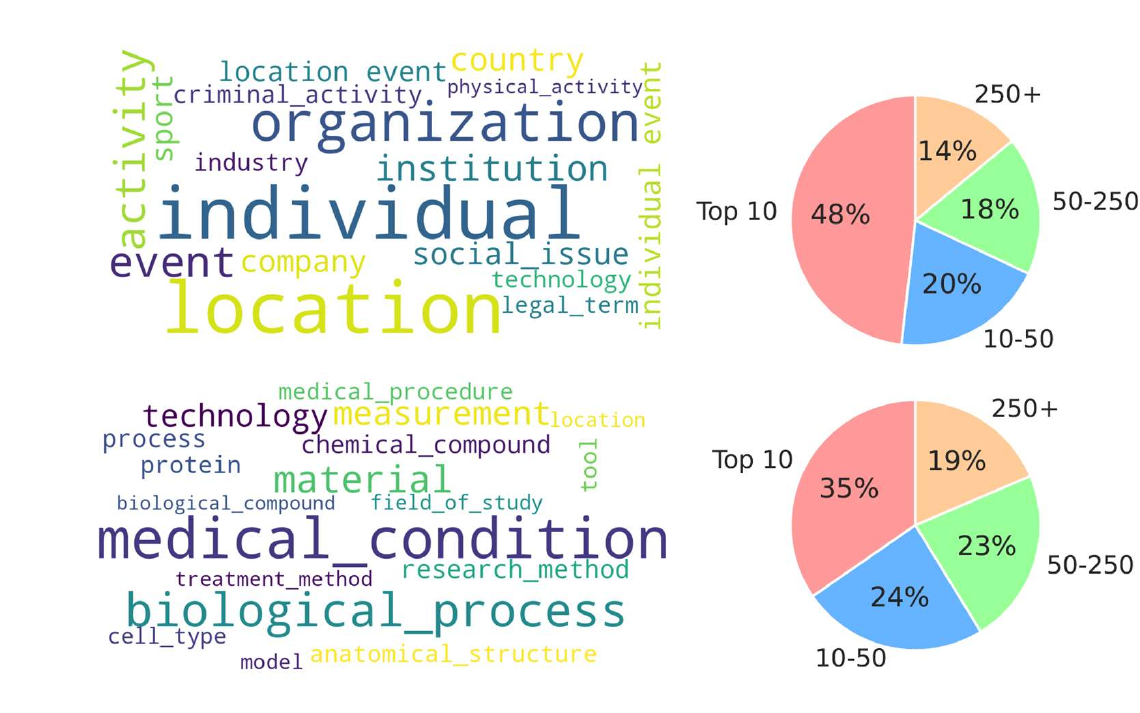}
\caption{A visualization of the goal distribution for the news domain (top) and the biomedical domain (bottom). \benchmarkname features both high-frequency goals and a diverse long-tail goal distribution.}
\label{metakp-stats-visualization}
\end{figure}

We curate a test set using each of these datasets and construct two domain-specific train/validation sets sampled from the training sets from KPTimes and KPBiomed. \cref{tab:dataset_stats} and \cref{metakp-stats-visualization} presents the basic statistics of the final datasets. Besides its domain coverage, one strength of \benchmarkname is its \textit{diverse} coverage: together, the dataset covers 3760 unique goals, including diverse topics and subjects. While 40\% of the instances correspond to the 10 most popular goals in each domain, a substantial number of goals also fall into the long tail distribution, posing significant new challenge in understanding the goal semantics. 

To construct \benchmarkname, the two-staged GPT-4 annotation costed approximately 500 USD, and the human annotators worked for approximately 80 hours in total on final data filtering. We randomly sample 50 documents each from KPTimes and KPBiomed, on which the annotators reach 0.699 Cohen's Kappa for inter-annotator agreement. Then, the annotators work on the rest documents separately. When ambiguous cases are found, a discussion is conducted to reach agreement. 

\paragraph{Irrelevant Goal Sampling} To test the ability of keyphrase generation models to abstain from generating keyphrases given irrelevant goals, for each document, we additionally construct a set of irrelevant goals. Concretely, we cluster the goals in the labelled data and use each document's existing goals as anchors to sample goals that are likely to be irrelevant to the document and thus it is unlikely that a keyphrase corresponding to the sampled goal exists for the document. We present the algorithm in the \cref{negative-sampling-details}. Using the algorithm, a balanced training set was created for training supervised methods for goal relevance judgment. 

\subsection{Evaluation Metric}

With \benchmarkname, we design two tasks to comprehensively evaluate a model's ability to perform on-demand keyphrase generation. 

\paragraph{Goal Relevance Assessment} This task aims to test whether a model can correctly distinguish irrelevant goals that cannot yield any keyphrase from the relevant goals. As we will show in \cref{metakp-for-event}, this skill is also crucial to enable a wide application of on-demand keyphrase generation models. Following recent literature on abstention \citep{feng2024don}, we use \textbf{Abstain F1} as the evaluation metric, which is defined as the harmonic mean of the precision and the recall of a model refusing to generate keyphrases for irrelevant goals. 

\paragraph{Goal-Oriented Keyphrase Generation} Given document $\mathcal{X}$, a list of goals $g_1, g_2, ..., g_n$, and references $\mathcal{Y}_{g_1}, \mathcal{Y}_{g_2}, ..., \mathcal{Y}_{g_n}$, we evaluate a model's predictions $P_1, P_2, ..., P_n$ with two metrics: 

\begin{compactenum}
    \item \textbf{Reference Agreement}, which assesses the model's ability to generate keyphrases specifically corresponding to the goal $g_i$. Concretely, we calculate and report $SemF1(Y_{g_i}, P_i)$, following \citet{wu2023kpeval}. 
    \item \textbf{Satisfactory Rate} ($SR$), which assesses the frequency of the model generating high-quality keyphrases. Concretely, we calculate and report $SR((\mathcal{Y}{g_1}, P_1), ..., (\mathcal{Y}{g_n}, P_n))$ as the percentage of goals that have $SemF1(Y_{g_i}, P_i)$ greater than a threshold\footnote{We fix $\tau=0.6$. This decision is based \citet{wu2023kpeval}, which suggests that the embedding model for $SemF1$ assigns a similarity score of approximately 0.6 for name variations.}. 
\end{compactenum}

%% file: tables/dataset_stats.tex
\begin{table}[t!]
    \setlength{\tabcolsep}{3.5pt}
    \centering
    \resizebox{\linewidth}{!} {%
    \begin{tabular}{c | c | c c c c c}
        \hline
        \textbf{Source} & \textbf{Split} & \textbf{\#Doc} & \textbf{\#Inst} & \textbf{\#Goal} & \textbf{|Goal|} & \textbf{\#KP/Goal} \\
        \hline
        \multirow{3}{*}{\textbf{KPTimes}} & Train & 1859 & 7502 & 1083 & 1.43 & 1.32 \\
         & Val & 100 & 392 & 148 & 1.46 & 1.37 \\
         & Test & 984 & 3836 & 679 & 1.41 & 1.33 \\
        \hdashline
        \textbf{DUC2001} & Test & 308 & 1642 & 549 & 1.50 & 1.53 \\
        \hdashline
        \multirow{3}{*}{\textbf{KPBiomed}} & Train & 1886 & 7807 & 1311 & 1.75 & 1.27 \\
        & Val & 100 & 404 & 189 & 1.75 & 1.32 \\
        & Test & 994 & 4136 & 865 & 1.76 & 1.27 \\
        \hdashline
        \textbf{Pubmed} & Test & 1269 & 4988 & 843 & 1.82 & 1.33 \\
        \hline
    \end{tabular}
    }
    \caption{Basic statistics of \benchmarkname. \#Inst = number of instances in the form ($\mathcal{X}, g, \mathcal{Y}_g$). |Goal| refers to the average number of words in $g$. Finally, \#KP/Goal corresponds to the average cardinality of $\mathcal{Y}_g$.}
    \label{tab:dataset_stats}
\end{table}

%% file: text/4_approach.tex
\section{Modeling Approach}
\label{section-approach}

In this section, we introduce two modeling approaches for on-demand keyphrase generation: a multi-task learning approach for fine-tuning sequence-to-sequence pre-trained language models, and a self-consistency decoding approach for prompting large language models (LLMs).

\subsection{Multi-Task Supervised Fine-tuning}
\label{section-approach-supervised}

Previous literature has demonstrated the effectiveness of fine-tuning sequence-to-sequence pre-trained language models for keyphrase generation \citep{kulkarni-etal-2022-learning, wu-etal-2022-representation, wu-etal-2023-rethinking-model}. However, it is unclear how these sequence prediction approaches could be adopted for on-demand keyphrase generation. To bridge this gap, we introduce a novel formulation to train a sequence-to-sequence model to autoregressively (1) assess the relevance of goals and (2) jointly consider the document as well as a desired goal to predict keyphrases. 

\begin{figure}[t!]
\centering
\includegraphics[width=0.98\linewidth]{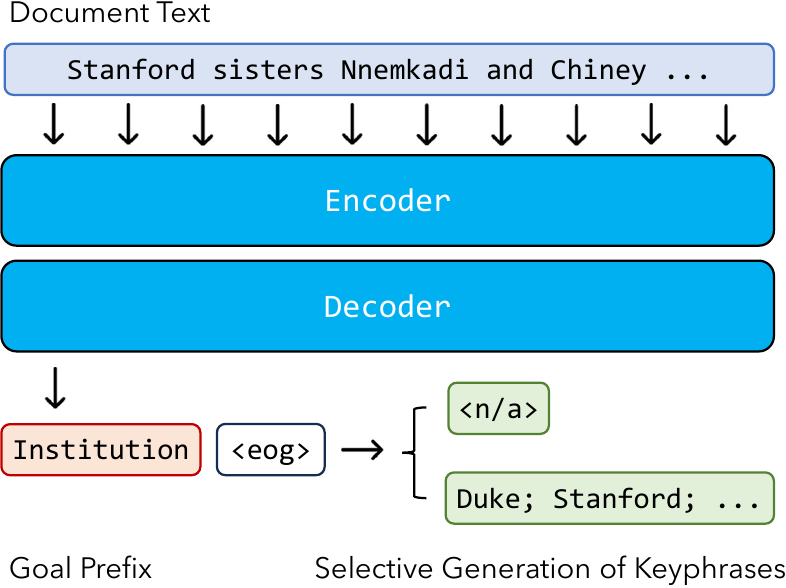}
\caption{A visualization of the inference process of the proposed sequence-to-sequence generation approach. Based on the document and the goal prefix, the model self-decides the relevance of the goal and selectively generates the keyphrases for relevant goals only.}
\label{seq2seq-arch-vis}
\end{figure}

Concretely, we formulate on-demand keyphrase generation as a hierarchical composition of two token prediction tasks. As shown in \cref{seq2seq-arch-vis}, with the document fed in the encoder, the decoder first models $P(g_i|\mathcal{X})$, the likelihood of $g_i$ being a high-quality relevant goal proposed by real users. The model verbalizes this probability in $P(\texttt{\textless n/a\textgreater}|\mathcal{X}, g_i)$, a special token for rejecting irrelevant goals. If the goal is determined as relevant, the model proceeds generating the keyphrases according to the distribution $P(\mathcal{Y}_{g_i}|\mathcal{X}, g_i)$ it learned. 

\paragraph{Inference} We use prefix-controlled decoding for inference. $g_i$, followed by a special end-to-goal token \texttt{\texttt{\textless eog\textgreater}}, is fixed as the decoder's start of generation. Then, we use autoregressive decoding to let the model self-assess the relevance of goal and automatically decide the keyphrases to generate. 

\paragraph{Training} We design a multi-task learning procedure to directly supervise the model on $P(\texttt{\textless n/a\textgreater}|\mathcal{X}, g_i)$ and $P(\mathcal{Y}_{g_i}|\mathcal{X}, g_i)$ with a mixture of relevant and irrelevant goals. As the goals provided by users could be arbitrary, we do not directly supervise the model on $P(g_i|\mathcal{X})$.

\paragraph{Remark} We note that the proposed approach has several advantages. First, both the goal relevance assessment and the keyphrase prediction process are streamlined in a single sequence prediction process, removing the need for separate architecture or inference pass. Second, since $g_i$ is not fed to the encoder, our model avoids the goal being diluted by the long input context and enables efficient inference by reusing the encoded input representation for predicting keyphrases with different goals.

\subsection{Prompting Large Language Models}
\label{section-approach-unsupervised}

Large language models (LLMs) that are tuned to follow human instructions have been shown to adapt well to a massive number of tasks defined through human queries \citep{ouyang2022training, OpenAI2023GPT4TR}. They have also been demonstrated to achieve promising keyphrase extraction or keyphrase generation performance, especially with semantic-based evaluation \citep{song2023chatgpt, wu2023kpeval}. As on-demand keyphrase extraction could be easily formulated as an instruction-following task, we investigate the potential of LLMs as an unsupervised approach. We start with a simple instruction for judging a goal's relevance:
\begin{center}
    \small
    \texttt{Decide if you should reject the high-level category given the title and abstract of a document. One could use the high-level category to write keyphrases from the document.}
\end{center}
as well as another instruction for keyphrase generation based on a goal: 
\begin{center}
    \small
    \texttt{Generate present and absent keyphrases belonging to the high-level category from the given text.}
\end{center}
Our preliminary experiments show that the first instruction already achieves a strong performance in deciding the goal relevance, even approaching supervised models (\cref{section-main-results}). However, when it comes to keyphrase generation, LLMs intriguingly misinterpret the task as named entity extraction: they often generate an almost exhaustive list of goal-related entities. To correct this behavior, we hypothesize that LLMs tend to generate salient entities more frequently and at an earlier location of the prediction sequence. Inspired by \citet{wang2023selfconsistency}, we thus design a novel self-consistency decoding process to leverage the rank and frequency information in LLMs' samples to filter out phrases that encode the most important information.

Concretely, using the same instruction and input, we sample $K$ prediction sequences $(s_1,...,s_K)$ from the LLM independently, each of which contains a variable number of keyphrases. Then, for each keyphrase $p$, we define its saliency score as:
\begin{equation}
\resizebox{0.9\hsize}{!}{$score(p)=\frac{freq(p)}{K}\times\frac{freq(p)}{{\sum_{i=1,...,K}rank(s_i, p)}}$}\nonumber,
\end{equation}
where $freq(p)$ returns the frequency of $p$ in all the samples and $rank(s_i, p)$ returns the rank of $p$ in $s_i$ (starting from 1) or 0 if $p\notin s_i$. The first term rewards keyphrases that frequently present in the samples, and the second term rewards keyphrases with a higher rank. Together, the score is defined to range from 0 to 1 regardless of the number of samples or the number of keyphrases a model generates per sample. Finally, we apply threshold filtering and only retain the high quality keyphrases with $score(p)$ greater than or equal to a threshold $\tau$. 

%% file: text/5_results.tex
\section{Experiments}
\label{section-experiments}

\begin{figure*}[ht!]
\centering
\includegraphics[width=0.95\textwidth]{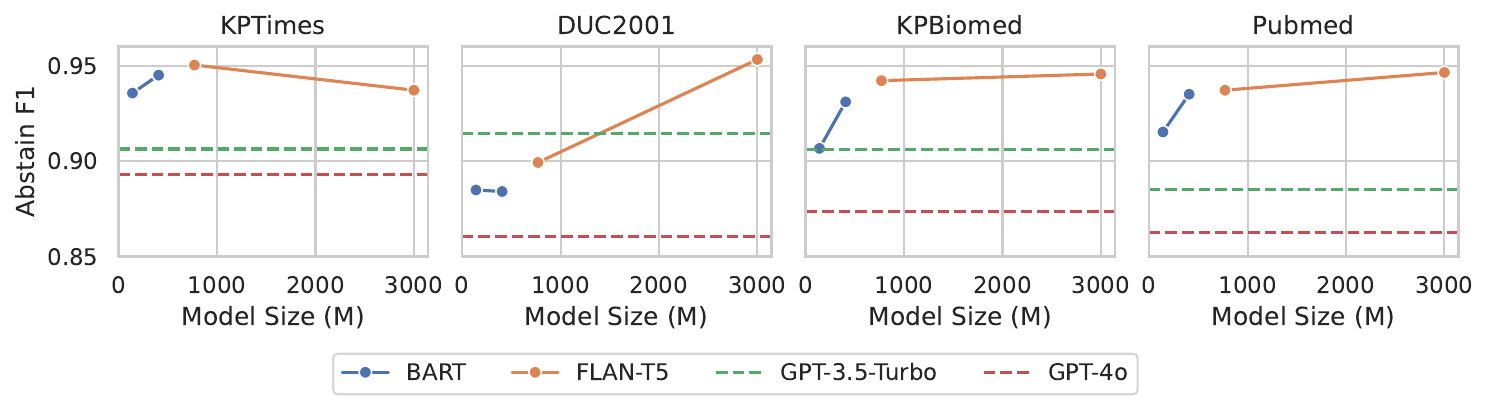}
\caption{Goal relevance judgment results of different types of models. Zero-shot prompting LLMs achieves a high performance, despite slightly falling below supervised models. Also, GPT-4o does not improve over GPT-3.5-Turbo.}
\label{rejection-results}
\end{figure*}

\input{tables/metakp_results_main}

\subsection{Experimental Setup}

\paragraph{Supervised Fine-tuning} Using the proposed objective, we fine-tune four sequence-to-sequence models: \texttt{BART-base/large} \citep{lewis-etal-2020-bart} and \texttt{Flan-T5-large/XL} \citep{pmlr-v202-longpre23a}, with diverse sizes ranging from 140M to 3B. We train the models for 20 epochs with batch size 64, learning rate 3e-5, the Adam optimizer, and a linear decay with 50 warmup steps. The best model checkpoint is chosen based on the keyphrase generation performance on the validation set.

\paragraph{Prompting} We use \texttt{gpt-3.5-turbo-0125} and the \texttt{gpt-4o-2024-05-13} models via the OpenAI API. We will denote the models as \texttt{GPT-3.5-Turbo} and \texttt{GPT-4o}. We use separate prompts for goal relevance judgment and on-demand keyphrase generation. For the first task, greedy search is used. For the second task, we generate 10 samples with temperature = 0.9. The output length is limited 30 tokens, which can accommodate approximately 10 keyphrases. Finally, for filtering, we use $\tau=0.3$ for all the datasets.

We document the full implementation details in \cref{appendix-impl-details}, including the prompt for language language models, the post-processing process, as well as the details for hyperparameter tuning. 

\subsection{Main Results}
\label{section-main-results}

We present the main results for the two tasks in \cref{rejection-results} and \cref{tab:main-results-kpgen}. 

\paragraph{Goal Relevance Assessment} According to \cref{rejection-results}, we find both supervised fine-tuning and unsupervised prompting reaches a high performance for assessing whether a goal, as indicated by over 0.85 Abstain F1 scores across all datasets. As model size scales, the out-of-distribution performance scales more readily, while the in-distribution performance plateaus at \texttt{Flan-T5-large}. With large language models, we observe strong performance especially on DUC2001, surpassing the performance of \texttt{Flan-T5-large} trained on KPTimes.

\paragraph{Keyphrase Generation} The main results for keyphrase generation are presented in \cref{tab:main-results-kpgen}. For supervised methods, we additionally include a "No Goal" baseline, where the model is fine-tuned to generate all the keyphrases for the same document at once. For both \texttt{BART-base} and \texttt{BART-large}, this baseline achieves a low performance, indicating the challenging nature of directly leveraging a keyphrase generation model for the proposed task. By comparison, the proposed goal-directed fine-tuning approach improves the performance by a large margin, with the best \texttt{Flan-T5-XL} model achieving 0.609 SemF1 and 0.544 satisfaction rate. On the other hand, directly zero-shot prompting large language models already achieves more superior performance compared to the supervised models trained without any goal. The proposed self-consistency further improves the performance substantially, allowing \texttt{GPT-4o} achieve 0.548 SemF1 and 0.475 satisfaction rate. Notably, results demonstrate that the LLM-based approach has the potential to be more generalizable. On DUC2001, all supervised models trained on KPTimes demonstrate a poor performance. By contrast, both \texttt{GPT-3.5-Turbo} and \texttt{GPT-4o} are able to surpass the performance of all supervised models.

In \cref{appendix-more-baselines}, we provide additional analyses on more relevant baselines: (1) reranking the outputs from supervised models by their similarity to the goal and (2) truncating LLMs' outputs to top-$k$ keyphrases. Results suggest that the proposed method still has more superior performance compared to these potentially stronger baselines.

\subsection{Analyses}
\label{section-analyses}

\begin{figure}[t!]
\vspace{-2mm}
\centering
\includegraphics[width=0.96\linewidth]{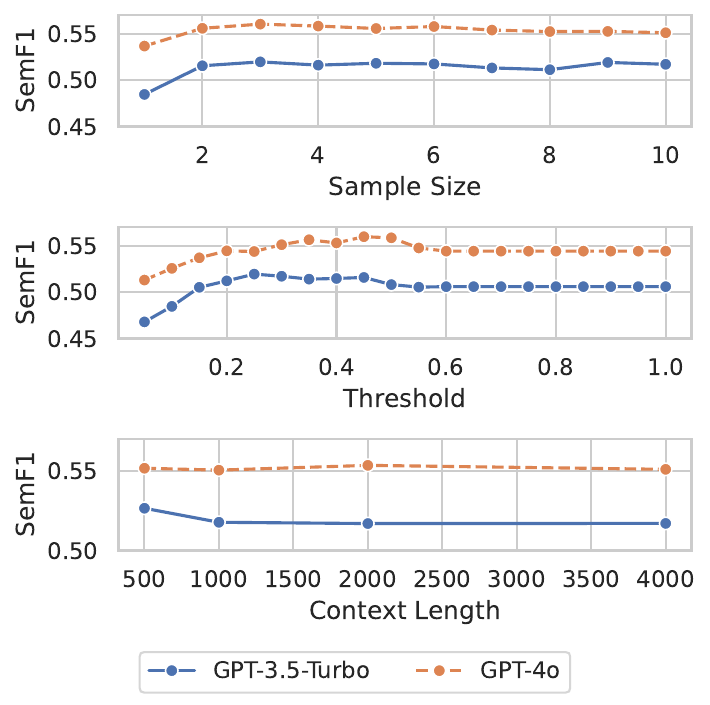}
\caption{Sensitivity of the self-consistency prompting approach's performance to number of samples, settings of threshold $\tau$, and the input length on KPTimes. The results on KPBiomed is presented in \cref{llm-hyperparameter-sensitivity-kpbiomed}.}
\label{llm-hyperparameter-sensitivity}
\vspace{-5mm}
\end{figure}

\paragraph{Which parameter affects LLMs the most?} In \cref{llm-hyperparameter-sensitivity}, we use use KPTimes' validation set to investigate the sensitivity of the LLM-based approach to three hyperparameters: number of samples ($K$), threshold $\tau$, and context length of the input. Although multiple samples are essential to high performance, more samples after two only help marginally. In addition, our method is insensitive to the threshold setting - the best performance can be obtained by multiple settings between 0.25 and 0.45. Finally, while \texttt{GPT-3.5-Turbo} exhibits a slight performance drop with longer context, \texttt{GPT-4o} is robust to context length variations. 

\paragraph{Does multi-task learning harm each individual task's performance?} 

In \cref{tab:loss-ablation}, we conduct an ablation study with \texttt{BART-base} on the supervised training loss. For each ablated component, we mask out the corresponding tokens when calculating the loss. Overall, combining the two learning objectives do not significantly harm the performance compared to only learning individual tasks, while incurring much less computational overhead. In fact, on KPTimes, the two tasks are constructive - learning goal relevance helps generating better goal-conforming keyphrases, and vice versa. 

\input{tables/loss_ablation}

%% file: tables/metakp_results_main.tex
\setlength{\tabcolsep}{3.5pt}
\begin{table*}[t!]
 \centering
 \resizebox{\linewidth}{!}{
 \begin{tabular}{c | c | c | c c | c c | c c | c c | c c }
 \hline
 \multirow{2}{*}{\textbf{Model}} & \multirow{2}{*}{\textbf{Size}} & \multirow{2}{*}{\textbf{Method}} & \multicolumn{2}{c|}{\textbf{KPTimes\textsuperscript{\ding{67}}}} & \multicolumn{2}{c|}{\textbf{DUC2001\textsuperscript{\ding{67}}}} &  \multicolumn{2}{c|}{\textbf{KPBiomed\textsuperscript{\ding{95}}}} & \multicolumn{2}{c|}{\textbf{Pubmed}\textsuperscript{\ding{95}}} & \multicolumn{2}{c}{\textbf{Average}}  \\
 & & &  \textbf{SemF1} & \textbf{SR} & \textbf{SemF1} & \textbf{SR} & \textbf{SemF1} & \textbf{SR} & \textbf{SemF1} & \textbf{SR} & \textbf{SemF1} & \textbf{SR}  \\
 \hline
 \hline
 \multicolumn{13}{c}{\textbf{\texttt{Supervised Methods}}} \\
 \hline
 \multirow{2}{*}{\texttt{BART-base}} & \multirow{2}{*}{140M} & No Goal & 0.395 & 0.192 & 0.299 & 0.089 & 0.300 & 0.107 & 0.305 & 0.196 & 0.325 & 0.146 \\
 & &  MetaKP & 0.728 & 0.699 & 0.447 & 0.319 & 0.508 & 0.417 & 0.504 & 0.406 & 0.547 & 0.460 \\
 \hdashline
 \multirow{2}{*}{\texttt{BART-large}} & \multirow{2}{*}{406M} & No Goal & 0.399 & 0.196 & 0.306 & 0.081 & 0.297 & 0.074 & 0.290 & 0.070 & 0.323 & 0.105 \\
 & &  MetaKP & 0.752 & 0.738 & 0.469 & 0.336 & 0.545 & 0.461 & 0.529 & 0.437 & 0.574 & 0.493 \\
 \hdashline
 \texttt{Flan-T5-large} & 770M &  MetaKP & \textbf{0.765} & \textbf{0.758} & 0.488 & 0.360 & 0.578 & 0.506 & 0.572 & 0.501 & 0.601 & 0.531 \\
 \texttt{Flan-T5-XL} & 3B &  MetaKP & 0.763 & 0.757 & 0.484 & 0.361 & \textbf{0.594}$^\dagger$ & \textbf{0.530}$^\dagger$ & \textbf{0.593}$^\dagger$ & \textbf{0.526}$^\dagger$ & \textbf{0.609}$^\dagger$ & \textbf{0.544}$^\dagger$ \\
  \hline\hline
 \multicolumn{13}{c}{\textbf{\texttt{Unsupervised Methods}}} \\
 \hline
 \multirow{2}{*}{\texttt{GPT-3.5-Turbo}} & \multirow{2}{*}{-} & Zero-Shot & 0.452 & 0.221 & 0.499 & 0.290 & 0.421 & 0.166 & 0.444 & 0.217 & 0.454 & 0.224 \\
 & &  Sample + SC & 0.518 & 0.406 & 0.572 & 0.516 & 0.513 & 0.423 & 0.472 & 0.376 & 0.519 & 0.430 \\
 \hdashline
 \multirow{2}{*}{\texttt{GPT-4o}} & \multirow{2}{*}{-} & Zero-Shot & 0.491 & 0.281 & 0.526 & 0.374 & 0.480 & 0.278 & 0.469 & 0.262 & 0.492 & 0.299 \\
 & &  Sample + SC & 0.552 & 0.460 & \textbf{0.578}$^\dagger$ & \textbf{0.535}$^\dagger$ & 0.529 & 0.451 & 0.532 & 0.453 & 0.548 & 0.475 \\
 \hline
\end{tabular}
 }
 \caption{Experiment results of supervised and unsupervised methods on-demand keyphrase generation. We use different superscripts to denote results that are reported using the models trained on KPTimes (\ding{67}) and KPBiomed (\ding{95}). SR = satisfaction rate. SC = self-consistency prompting The best results are boldfaced. $^\dagger$statistically significantly better than the second highest result with $p<0.01$, tested via paired t-test.}
 \label{tab:main-results-kpgen}
\end{table*}

%% file: tables/loss_ablation.tex
\setlength{\tabcolsep}{3.5pt}
\begin{table}[!t]
    \small
    \centering
    \resizebox{0.9\linewidth}{!} {%
    \begin{tabular}{l | c c | c c }
    \hline
    \multirow{2}{*}{\textbf{Objective}}  & \multicolumn{2}{c|}{\textbf{ID}} & \multicolumn{2}{c}{\textbf{OOD}} \\
    & \textbf{AF1} & \textbf{SR} & \textbf{AF1} & \textbf{SR} \\
    \hline
    \multicolumn{5}{c}{\textbf{\texttt{Training on KPTimes}}} \\
    \hdashline
    Multi-task Learning & \textbf{0.936} & \textbf{0.699} & 0.885 & \textbf{0.319} \\
    Goal Relevance Only & 0.928 & - & \textbf{0.898} & - \\
    Keyphrase Only & - & 0.692 & - & 0.316  \\
    \hline
    \multicolumn{5}{c}{\textbf{\texttt{Training on KPBiomed}}} \\
    \hdashline
    Multi-task Learning & 0.907 & 0.417 & 0.915 & 0.406  \\
    Goal Relevance Only & \textbf{0.917} & - & \textbf{0.916} & - \\
    Keyphrase Only & - & \textbf{0.425}  &  - & \textbf{0.407} \\
    
    \hline
    \end{tabular}
    }
    \caption{Ablation study on the multi-task learning setup. AF1 = Abstain F1, SR = Satisfaction Rate.
    }
    \label{tab:loss-ablation}
\end{table}

%% file: text/6_discussion.tex
\section{Discussion}

In this section, we further discuss the real-world use cases of the on-demand keyphrase generation. Then, we apply it to solve event detection as a proof-of-concept case study.

\subsection{Applications}

We identify five crucial applications of on-demand keyphrase generation:

\paragraph{Programmatic topic analysis} Given a set of topics, topic analyzers such as news editors, website managers, or policymakers can utilize an on-demand keyphrase generation model to sweep across a set of documents and attempt to extract keyphrases with the topics as the goals. In this way, the topic distribution could be obtained, along with the supporting keyphrases. If the algorithm is applied to sets of documents across time, the evolution of the topic distribution could be obtained. We demonstrate this application in the next section.

\paragraph{Ontology-guided keyphrase annotation} In practice, there are generally guidelines for human annotators on how to write good keyphrases. A common requirement is to write keyphrases only belonging to specific types. In such cases, using an on-demand keyphrase generation model in the loop to propose keyphrase candidates for specific types can improve both the efficiency of keyphrase annotation and the consistency between different human annotators.

\paragraph{Keyphrase classification and visualization} An on-demand keyphrase generation model can be automatically repurposed for keyphrase category classification. Such classification could be conducted programmatically for a set of predefined keyphrase types. The results could also be used for further visualization for the downstream user.

\paragraph{Multi-view indexing} With an on-demand keyphrase generation model, it is possible to generate sets of keyphrases focusing on different categories. These keyphrases could be used in the index as different views of the same document. This design would improve the retrieval performance especially with dense retrievers, as jointly encoding a large number of unrelated keyphrases could harm the quality of the resulting embedding.

\paragraph{Keyphrase expansion}
On-demand keyphrase generation allows the user to expand output of existing keyphrase systems by prompting the model with the desired goal. This is a common use case in the news domain, where editors need to understand and compare the topic of one or more articles. 

\subsection{\textsc{MetaKP} in the Wild: Event Detection}
\label{metakp-for-event}

Finally, we demonstrate the potential of on-demand keyphrase generation as general NLP infrastructure, using event detection (ED) as a case study. 

We leverage the testing dataset used in SPEED \citep{parekh2024event}, which contains time-stamped social media posts related to Monkeypox\footnote{We solicited the dataset and outputs from the authors.}. From the SPEED ontology, we curate seven epidemic-related goals: \textit{disease infection, epidemic spread, epidemic prevention, epidemic control, symptom, recover from disease, death from epidemic}. Then, we run a \texttt{FLAN-T5-large} model trained on all training data from \benchmarkname to assess the relevance of each goal against each social media post. If the probability of \texttt{\texttt{\textless n/a\textgreater}} following \texttt{\texttt{\textless eog\textgreater}} is smaller than 0.001, the goal is judged relevant to the post and thus the underlying event is likely. 

\begin{figure}[t!]
\centering
\includegraphics[width=0.95\linewidth]{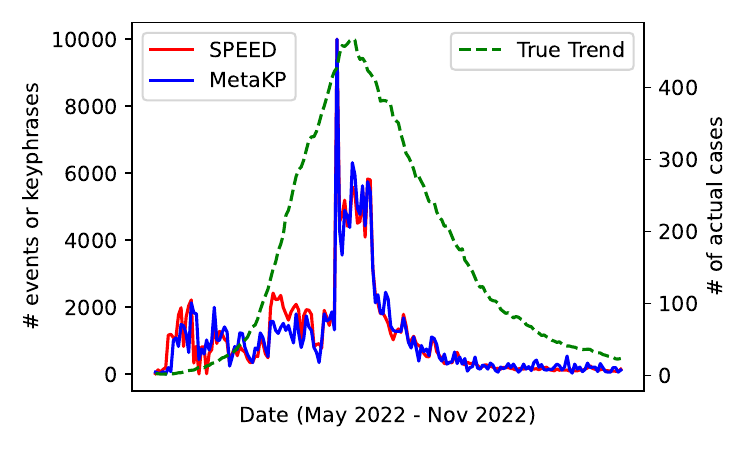}
\caption{Number of events/keyphrases extracted for Monkeypox as a function of time. The true trend and SPEED outputs are solicited from \citet{parekh2024event}.}
\label{metakp-for-ee-results}
\end{figure}

As shown in \cref{metakp-for-ee-results}, we observe that this keyphrase-based paradigm is able to extract trends that are similar to an ED model trained on SPEED \citep{parekh2024event}. Intuitively, given a sentence containing "getting vaccination", instead of focusing on the trigger "get", on-demand keyphrase generation is able to focus more on "vaccination", given the goal "epidemic control". In this way, on-demand keyphrase generation models can both be naturally repurposed for ED and also promises to extract supporting topics related to the the event.

%% file: text/7_conclusion.tex
\section{Conclusion}

We introduce on-demand keyphrase generation, targeting the need for dynamic, goal-oriented keyphrase prediction tailored to diverse applications and user demands. A large-scale, multi-domain, human-verified benchmark \benchmarkname was curated and introduced. We designed and evaluated both supervised and unsupervised methods on \benchmarkname, highlighting the strengths of self-consistency prompting with large language models. This approach significantly outperformed traditional fine-tuning methods under domain shifts, showcasing its robustness and the broader applicability of our methodology. Finally, we underscore the versatility of on-demand keyphrase generation in practical applications such as epidemic event extraction, promising a new direction for keyphrase generation as general NLP infrastructure.

\section*{Limitations}

In this work, we propose the novel on-demand keyphrase generation paradigm. In the future, several exciting directions exist for extending the paradigm as well as the \benchmarkname benchmark:

\begin{compactenum}
    \item \textbf{Multi-lingual Keyphrase Generation}. \benchmarkname only covers data in English. Further benchmarking and enhancing the multilingual and cross-lingual on-demand keyphrase generation ability is an important future direction. 
    \item \textbf{Wider Domain Coverage}. We mainly focus on the news and the biomedical text domain as they have been shown as important application domains for keyphrase generation. 
    \item \textbf{Flexible Instructions}. In this work, the "demand" from the users are generally defined as topics or categories of keyphrases. However, future work could expand this definition to include demands that specify stylistic constraints such as the number of keyphrases, the length, and their formality. 
\end{compactenum}

\section*{Ethics Statement}

As a new task and paradigm, on-demand keyphrase generation may bring new security risks and ethical concerns. To begin with, although keyphrase generation models generally have outstanding understanding of phrase saliency, they generally have a shallower understanding of semantics and factuality. Thus, when pairing keyphrases with goals, potential misinformation could be created. For instance, when queried with "cure" as a goal, a model may return certain concepts that are factually wrong. In addition, when queries contain certain occupations as goals, a keyphrase generation model may reinforce existing gender stereotypes by selectively generating and ignoring entities with a certain gender. We view these possibilities as potential risks and encourage a thorough redteaming process before deploying on-demand keyphrase generation systems in real-world products.

We use KPTimes and KPBiomed data distributed by the original authors. For DUC2001 and PubMed, we access the data via ake-datasets\footnote{\url{https://github.com/boudinfl/ake-datasets}}. KPTimes was released under Apache-2.0 license, and we cannot find licensing information for DUC2001, KPBiomed, and PubMed. ake-datasets was also released under Apache-2.0. No additional preprocessing is performed in \benchmarkname except lower-casing and tokenization. While we mainly rely on the original authors for dataset screening to remove sensitive and harmful information, we also actively monitor the data quality during in the human filtering process and remove any document that could cause privacy or ethics concerns. As OpenAI models are involved in the data curation process, our code and datasets will be released with MIT license with a research-only use permission. 

\section*{Acknowledgments}
The research is supported in part by Taboola. We thank the Taboola team for the helpful discussion. We also thank Da Yin, Tanmay Parekh, Zongyu Lin, and other members of the UCLA-NLP group for their valuable feedback.

%% file: text/appendix.tex
\appendix

\clearpage
\appendix
\twocolumn[{%
 \centering
 \Large\bf Supplementary Material: Appendices \\ [20pt]
}]

\section{\benchmarkname Construction Details}
\label{appendix-dataset-details}

In this section, we describe the details of the construction process of \benchmarkname. 

\subsection{GPT-4 Annotation}

\paragraph{Goal Proposal} In \cref{goal-proposal-prompt}, we show the prompt used to instruct GPT-4 to propose goals from the document and human-annotated keyphrases. We truncate the document body to four sentences as its role is only providing essential contextual. The LLM is instructed to propose all the goals for all the keyphrases together, which helps the model group together keyphrases that share the same goal.

\begin{figure}[ht!]
    \centering
    \small
    \fbox{\begin{tabular}{ p{0.94\linewidth} }
    \texttt{Document Title: \{title\}} \\
    \texttt{First 4 sentences of the document body: \{body\}} \\
    \\
    \texttt{Keyphrases (separated by ";"): \{keyphrases\}}
    \\
    \\
    \texttt{For each keyphrase, generate an abstract category for the keyphrase. Examples include process, task, material, tool, measurement, model, technology, and metric etc. Do not limit yourself to the examples. Make sure that the categories are informative in the domain of science and appearing natural as if that assigned by a well-read user. Return a list of dictionaries, each with two keys - "keyphrase" and "category". If two keyphrases have the same category, make sure that they are labelled with the same phrase. Do not change how the keyphrases appear, including their cases. Return json only and do not say anything else.
    }

    \end{tabular}}
    \caption{Prompt used for instructing GPT-4 to generate the goals from a document and keyphrases.}
    \label{goal-proposal-prompt}
    \vspace{-2mm}
\end{figure}

\paragraph{Goal Refinement} Then, we instruct GPT-4 to refine the goals by trying to generate more abstract versions of them. The prompt is shown in \cref{goal-refinement-prompt}. As we perform the refinement directly from the chat history of the previous step, we omit the previous prompt and step 1 model outputs.

\begin{figure}[ht!]
    \centering
    \small
    \fbox{\begin{tabular}{ p{0.94\linewidth} }
    \texttt{... step 1 prompt and model outputs ...} \\
    \\
    \texttt{Can you make the categories more abstract, yet still informative to the keyphrase? If the categories are already abstract enough, you do not need to change. Return json only.}

    \end{tabular}}
    \caption{Prompt used for instructing GPT-4 to improve the abstractiveness of the proposed goals.}
    \label{goal-refinement-prompt}
    \vspace{-2mm}
\end{figure}

For both of the steps, we use greedy search and cap the output to 400 tokens. We parse the results string into json format to extract the goals.

\subsection{Human Validation}

Next, based on the two rounds of proposed goals, the two authors (student researchers familiar with NLP and the keyphrase generation task) filter out high quality goals as the final benchmarking dataset. We emphasize that this decision is required due to the nature of the task, which requires expert annotators to ensure a high data quality. The consent to use and release the annotation traces was obtained from both of the authors. The type of research conducted by this work is automatically determined exempt from by the authors' institution's ethics review board. We design and enforce two major guidelines during the annotation process:

\begin{compactenum}
    \item Remove a goal if it is semantically equivalent to or a subtype of some another goal that is more abstract.
    \item Remove a goal if it so abstract that it could also enclose other keyphrases not currently paired with the goal. This criterion includes overly vague goals (e.g., "concept") and goals that corresponds to the topic of the entire pssage (e.g., "chemistry concepts"). 
\end{compactenum}

As mentioned in \cref{section-benchmark}, this process allows the annotator reach a high inter-annotator agreement of 0.699 Cohen's Kappa. In addition, the annotatos actively engage in a discussion whenever ambiguous cases are found. Finally, we conduct a rule-based postprocessing with two stages. 

\begin{compactenum}
    \item \textbf{Goal Removal}. We remove the following goals as they represent overly general goals: \texttt{entity, process, concept}.
    \item \textbf{Goal Unification}. We merge the following goal labels as they represent the same meaning. \cref{tab:goal-merging} presents the source and target goals. Note that to preserve the diversity of the goals, we refrain from merging aggressively and only merge the basic cases that may be result from annotation discrepancy. 
\end{compactenum}

\begin{table}[h!]
    \setlength{\tabcolsep}{3.5pt}
    \centering
    \resizebox{\linewidth}{!} {%
    \begin{tabular}{c c}
        \hline
        \textbf{Source Goals} & \textbf{Target Goals} \\
        place, geographical location & location \\
        person, people, individual person & individual \\
        geopolitical entity & country \\
        ... event & event \\
        profession & occupation \\
        belief system & religion \\
        incident outcome & outcome \\
        subject & topic \\
        incident & event \\
        ... equipment & equipment \\
        ... procedure & procedure \\
        \hline
    \end{tabular}
    }
    \caption{Goal merging directions for \benchmarkname label cleaning. We replace all occurrences of source goals with target goals. }
    \label{tab:goal-merging}
\end{table}

\subsection{Negative sampling Algorithm}
\label{negative-sampling-details}

To construct the training and evaluation data for evaluating the model's ability to reject irrelevant goals, we design a simple algorithm to sample irrelevant goals. Concretely, we pool together all the existing goals from the same dataset as the universal goal set and leverage the phrase embedding model released by \citep{wu2023kpeval} to embed all the phrases. Then, for each goal from the document, we use it as an anchor to retrieve $d$\% most dissimilar goals. We use $d=50$ for all the datasets. From these goals, we sample a goal that is not associated with the document as the irrelevant goal according to the frequency distribution of these goals appearing as relevant goals in the final dataset. We additionally design a frequency match constraint, which enforces that the frequency of a goal $g$ appearing as an irrelevant goal should not exceed the frequency it appears as a relevant goal. In practice, the frequency match constraint is applied first. If no eligible goals remain, we sample a goal from the $d$\% most dissimilar goals according to frequency.

\section{Implementation Details}
\label{appendix-impl-details}
\label{appendix-hyperparameter}

\subsection{Supervised Fine-tuning}
\label{apprendix-supervised-ft}

For multi-task learning with \texttt{BART} and \texttt{Flan-T5}, we base our implementation on the Huggingface Transformers implementations provided by \citep{wu-etal-2023-rethinking-model} and train for 20 epochs with early stopping. We use learning rate 3e-5, linear decay, batch size 64, and the AdamW optimizer. Due to the context limitations of \texttt{Flan-T5}, all the input documents for \texttt{BART} and \texttt{Flan-T5} are truncated to 512 tokens to enable a fair comparison. We perform a careful hyperparameter search over the learning rate, batch size, and warm-up steps. The corresponding search spaces are \{1e-5, 3e-5, 6e-5, 1e-4\}, \{16, 32, 64, 128\}, and \{50, 100, 250, 500\}. The best hyperparameters are chosen based on the performance on the validation set. To decode from the fine-tuned models, we fix the decoder's prefix using the constrained decoding functionalities provided by Huggingface Transformers and use greedy search to complete the suffix. 

The fine-tuning experiments are performed on a local GPU server with eight Nvidia RTX A6000 GPUs (48G each). We use gradient accumulation to achieve the desired batch sizes. Fine-tuning \texttt{BART-base}, \texttt{BART-base}, \texttt{Flan-T5-large}, and \texttt{Flan-T5-XL} take, respectively. 


\subsection{Large Language Models}
\label{appendix-llm-prompt}

We present the prompts for prompting large language models for goal relevance judgment and goal-conforming keyphrase generation in \cref{relevance-judgment-prompt} and \cref{keyphrase-gen-prompt}.

\begin{figure}[ht!]
    \centering
    \small
    \fbox{\begin{tabular}{ p{0.94\linewidth} }
    \texttt{In this task you will need to decide if you should reject the high-level category given the title and abstract of a document. One could use the high-level category to write keyphrases from the document. If you decide the category is relevant to the document, generate yes; if the category is not relevant, generate no. Do not output anything else.} \\
    \\
    \texttt{Document Title: \{title\}} \\
    \texttt{Document Abstract: \{body\}} \\
    \\
    \texttt{High-level Category: \{goal\}} \\
    \texttt{Relevant? (yes or no):}
    \end{tabular}}
    \caption{Prompt used for goal relevance judgment.}
    \label{relevance-judgment-prompt}
    \vspace{-2mm}
\end{figure}

\begin{figure}[ht!]
    \centering
    \small
    \fbox{\begin{tabular}{ p{0.94\linewidth} }
    \texttt{Generate present and absent keyphrases belonging to the high-level category from the given text, separated by commas. Do not output anything else.} \\
    \\
    \texttt{Document Title: \{title\}} \\
    \texttt{Document Abstract: \{body\}} \\
    \\
    \texttt{High-level Category: \{goal\}} \\
    \texttt{Keyphrases (Must be of category "\{goal\}"):}
    \end{tabular}}
    \caption{Prompt used for on-demand keyphrase generation with LLMs.}
    \label{keyphrase-gen-prompt}
    \vspace{-2mm}
\end{figure}

For all the results reported in the paper, we use \texttt{gpt-3.5-turbo-0125} and the \texttt{gpt-4o-2024-05-13} models via the OpenAI API. 

For goal relevance judgment, we use greedy decoding and record the yes/no predictions for evaluation. The document body is truncated to the first five sentences as we find providing longer context barely improves the performance. 

\begin{figure}[ht!]
\vspace{-2mm}
\centering
\includegraphics[width=0.9\linewidth]{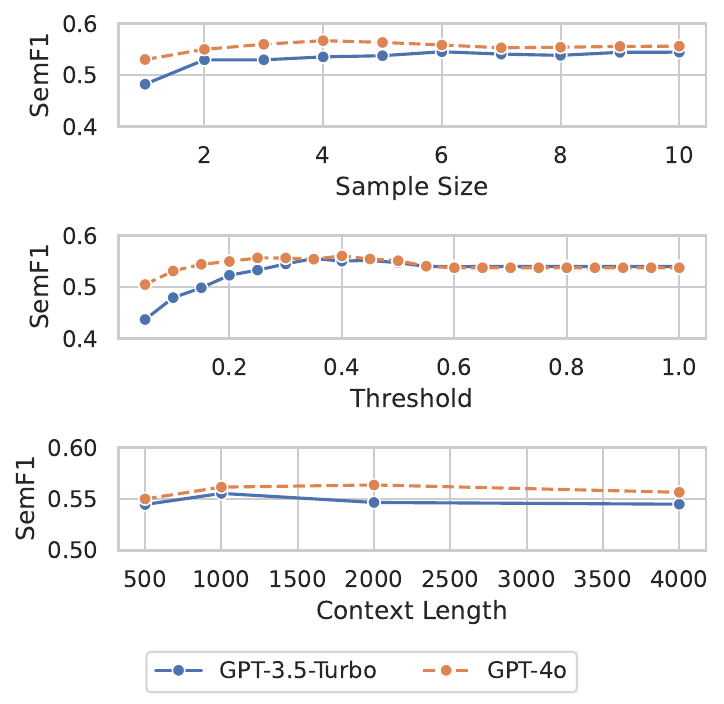}
\caption{Sensitivity of the self-consistency prompting approach's performance to number of samples, settings of threshold $\tau$, and the input length on KPBiomed.}
\label{llm-hyperparameter-sensitivity-kpbiomed}
\end{figure}

For on-demand keyphrase generation, the input length is truncated to 4000 tokens. We generate 10 samples with temperature = 0.9. The output length is limited to 30 tokens, which accommodate approximately 10 keyphrases. Finally, for filtering, we set a fixed threshold $\tau=0.3$. We lower-case all the outputs and use a string matching algorithm to remove excessive parts generated by the model such as "present keyphrases: ". The method's sensitivity to the hyperparameter settings is presented in \cref{llm-hyperparameter-sensitivity} and \cref{llm-hyperparameter-sensitivity-kpbiomed}.

Since the proposed LLM-based methods are unsupervised, we refrain from extensively tuning the hyperparameters. The only exception is that we use the validation sets to determine a reasonable good setting of the sample size $K$ and the threshold $\tau$, which is uniformly applied to all the datasets.

\section{Further Analyses}

\subsection{Additional Baselines}
\label{appendix-more-baselines}

In this section, we compare the proposed methods with two more relevant baselines. First, we compare fine-tuning with the proposed objective with directly reranking the outputs from the ``No Goal" model. we use the phrase embedding model in \citep{wu2023kpeval} to encode the goal and each of the “No Goal” model’s predicted keyphrases and then keep the top-k most similar keyphrases. Second, we compare the proposed prompting strategy with directly truncating the zero-shot outputs to top-k keyphrases. As shown in \cref{tab:more-baselines-kpgen}, the reranking baseline still has a large gap compared with the proposed fine-tuning objective, even if the two are trained on the same set of documents and keyphrases. For truncation, we find that it could lead to a higher performance for most of the datasets, but still underperforms the proposed method. Overall, these results still suggest the superiority of the proposed methods.

\input{tables/more_baselines}

\subsection{Prompt Sensitivity}
\label{appendix-prompt-sensitivity}

Inspired by \citep{DBLP:journals/corr/abs-2405-16571}, we conduct an analysis of our LLM-based method's sensitivity to the variation of the prompt. Our original instruction is "\texttt{Generate present and absent keyphrases belonging to the high-level category from the given text, separated by commas. Do not output anything else}", and we evaluate four edits:
\begin{compactenum}
    \item  “present and absent keyphrases” -> “all keyphrases”
    \item “keyphrases” -> “keywords”
    \item “generate” -> “extract”
    \item “category” -> “type”
\end{compactenum}

\input{tables/prompt_sensitivity}

We show the performance of GPT-3.5-Turbo on the validation sets in \cref{prompt-sensitivity-results}. Although a fluctuation could be observed under prompt perturbation, its magnitude is not large, indiciating the relative robustness of LLM-based approach to different prompt styles.

\section{Qualitative Study}

\input{tables/qualitative_study}

In \cref{qualitative-study-news} and \cref{qualitative-study-biomed}, we present and compare the outputs of \texttt{Flan-T5-XL}, zero-shot sampling from \texttt{GPT-4o}, and self-consistency sampling from \texttt{GPT-4o} in two domains. Compared to supervised models, which often generates suboptimal keyphrases under distribution shift, GPT-4o exhibits consistent high recall across domains, and the self-consistency reranking process further filters high quality goals from the zero-shot keyphrase predictions across multiple samples.

%% file: tables/more_baselines.tex
\setlength{\tabcolsep}{3.5pt}
\begin{table*}[t!]
 \centering
 \resizebox{\linewidth}{!}{
 \begin{tabular}{c | c | c | c c | c c | c c | c c | c c }
 \hline
 \multirow{2}{*}{\textbf{Model}} & \multirow{2}{*}{\textbf{Size}} & \multirow{2}{*}{\textbf{Method}} & \multicolumn{2}{c|}{\textbf{KPTimes\textsuperscript{\ding{67}}}} & \multicolumn{2}{c|}{\textbf{DUC2001\textsuperscript{\ding{67}}}} &  \multicolumn{2}{c|}{\textbf{KPBiomed\textsuperscript{\ding{95}}}} & \multicolumn{2}{c|}{\textbf{Pubmed}\textsuperscript{\ding{95}}} & \multicolumn{2}{c}{\textbf{Average}}  \\
 & & &  \textbf{SemF1} & \textbf{SR} & \textbf{SemF1} & \textbf{SR} & \textbf{SemF1} & \textbf{SR} & \textbf{SemF1} & \textbf{SR} & \textbf{SemF1} & \textbf{SR}  \\
 \hline
 \hline
 \multicolumn{13}{c}{\textbf{\texttt{Supervised Methods}}} \\
 \hline
 \multirow{5}{*}{\texttt{BART-base}} & \multirow{5}{*}{140M} & No Goal & 0.395 & 0.192 & 0.299 & 0.089 & 0.300 & 0.107 & 0.305 & 0.196 & 0.325 & 0.146 \\
 & & No Goal + Rerank (top-1) & 0.501 & 0.377 & 0.336 & 0.185 & 0.348 & 0.231 & 0.334 & 0.199 & 0.341 & 0.209 \\
 & & No Goal + Rerank (top-2) & 0.491 & 0.403 & 0.348 & 0.146 & 0.366 & 0.236 & 0.356 & 0.213 & 0.345 & 0.199 \\
 & & No Goal + Rerank (top-3) & 0.462 & 0.299 & 0.337 & 0.091 & 0.360 & 0.153 & 0.354 & 0.160 & 0.337 & 0.154 \\
 & & MetaKP & \textbf{0.728} & \textbf{0.699} & \textbf{0.447} & \textbf{0.319} & \textbf{0.508} & \textbf{0.417} & \textbf{0.504} & \textbf{0.406} & \textbf{0.547} & \textbf{0.460} \\
  \hline\hline
 \multicolumn{13}{c}{\textbf{\texttt{Unsupervised Methods}}} \\
 \hline
 \multirow{5}{*}{\texttt{GPT-3.5-Turbo}} & \multirow{5}{*}{-} & Zero-Shot & 0.452 & 0.221 & 0.499 & 0.290 & 0.421 & 0.166 & 0.444 & 0.217 & 0.454 & 0.224 \\
 & & Zero-Shot + Trunc (top-1) & 0.490 & 0.372 & 0.545 & 0.448 & 0.507 & 0.416 & 0.420 & 0.315 & 0.491 & 0.388\\
 & & Zero-Shot + Trunc (top-2) & 0.506 & 0.396 & 0.558 & \textbf{0.518} & 0.493 & 0.402 & 0.465 & 0.367 & 0.506 & 0.421 \\
 & & Zero-Shot + Trunc (top-3) & 0.493 & 0.342 & 0.547 & 0.474 & 0.469 & 0.302 & 0.470 & 0.330 & 0.495 & 0.362 \\
 & &  Sample + SC & \textbf{0.518} & \textbf{0.406} & \textbf{0.572} & 0.516 & \textbf{0.513} & \textbf{0.423} & \textbf{0.472} & \textbf{0.376} & \textbf{0.519} & \textbf{0.430} \\
 \hline
\end{tabular}
 }
 \caption{Comparison with additional supervised and unsupervised baselines on MetaKP. The proposed methods still outperform these more relevant baselines in both supervised and unsupervised settings.}
 \label{tab:more-baselines-kpgen}
\end{table*}

%% file: tables/prompt_sensitivity.tex
\begin{table*}[h!]
\centering
\begin{tabular}{l|cc|cc}
\hline
\textbf{Prompt} & \multicolumn{2}{c|}{\textbf{KPTimes (validation)}} & \multicolumn{2}{c}{\textbf{KPBioMed (validation)}} \\ 
 & \textbf{Sem F1} & \textbf{SR} & \textbf{Sem F1} & \textbf{SR} \\ \hline
Original prompt & 0.449 & 0.204 & 0.440 & 0.220 \\ 
Edit 1 & 0.462 & 0.217 & 0.456 & 0.220 \\ 
Edit 2 & 0.427 & 0.189 & 0.406 & 0.146 \\ 
Edit 3 & 0.437 & 0.176 & 0.429 & 0.193 \\ 
Edit 4 & 0.435 & 0.225 & 0.451 & 0.220 \\ \hline
\end{tabular}
\caption{Validation results for different prompt edits on KPTimes and KPBioMed datasets.}
\label{prompt-sensitivity-results}
\end{table*}

%% file: tables/qualitative_study.tex
\begin{figure*}[h!]
\small
\centering
\begin{tabular}{p{0.98\linewidth}}
    \hline
    \multicolumn{1}{c}{\texttt{Domain: News}} \\
    \hline
    \hline
    \underline{\texttt{Source: KPTimes}} \\
    \textbf{Title: } \textnormal{more connecticut schools fall short of ``no child left behind'' standards}  \\
    \textbf{Abstract: } \textnormal{since [digit] , when president bush signed the federal no child left behind law , schools in connecticut have scrambled to revamp curriculums , step up professional development for teachers and continually assess students ’ test scores to comply with the law ’ s requirements . it is an effort that has dominated the agendas of school officials not just in connecticut , but all over the nation , and not everyone is happy that test results have become such a focus . when results of the latest test scores were announced a few weeks ago , about [digit] percent , or [digit] , of the public schools in connecticut did not make the grade under the federal law , state officials said . the state added [digit] schools to its list of schools that failed to meet the federal benchmarks . }  \\
    \hdashline
    \textcolor{blue}{\textbf{Goal 1: } \textnormal{institution}}  \\
    \textcolor{teal}{\textbf{Reference:} \textnormal{education and schools}} \\
    \textbf{Prediction (\texttt{Flan-T5-XL}): } \textnormal{education and schools}  \\
    \textbf{Prediction (\texttt{GPT-4o (zero-shot}): } \textnormal{connecticut schools, federal government, state department of education, norwalk public schools, greenwich school district, greenwich public schools}  \\
    \textbf{Prediction (\texttt{GPT-4o (self-consistency}):}  \textnormal{connecticut schools, state department of education} \\
    \hdashline
    \textcolor{blue}{\textbf{Goal 2:}  \textnormal{law}} \\
    \textcolor{teal}{\textbf{Reference:} \textnormal{no child left behind act}} \\
    \textbf{Prediction (\texttt{Flan-T5-XL}): } \textnormal{no child left behind act}  \\
    \textbf{Prediction (\texttt{GPT-4o (zero-shot}): } \textnormal{no child left behind, federal no child left behind law, federal law, federal government, new accountability system, adequate yearly progress}  \\
    \textbf{Prediction (\texttt{GPT-4o (self-consistency}):} \textnormal{no child left behind, no child left behind law} \\
    \hline
    \hline
    \underline{\texttt{Source: DUC2001 (out-of-distribution)}} \\
    \textbf{Title:} (empty) \\
    \textbf{Abstract: } \textnormal{millions of gallons of crude oil that spilled when a tanker ran aground spread across a wildlife - rich stretch of ocean saturday , and alaska ' s chief environmental officer criticized cleanup efforts as too slow . the biggest oil spill in u . s . history created a slick about seven miles long and seven miles wide in prince william sound . the coast guard said only reef island and the western edge of bligh island had been touched by the slick . `` this situation , i think , was everyone ' s secret nightmare about what could happen with oil traffic in the sound ,'' said dennis kelso , commissioner of the alaska department of environmental conservation .}  \\
    \hdashline
    \textcolor{blue}{\textbf{Goal 1: } \textnormal{substance}}  \\
    \textcolor{teal}{\textbf{Reference:} \textnormal{crude oil}} \\
    \textbf{Prediction (\texttt{Flan-T5-XL}): } \textnormal{oil ( petroleum ) and gasoline}  \\
    \textbf{Prediction (\texttt{GPT-4o (zero-shot}): } \textnormal{crude oil, oil spill, oil pollution, north slope crude oil, spilled oil, leaking oil, oil slick, spilled crude oil}  \\
    \textbf{Prediction (\texttt{GPT-4o (self-consistency}):} \textnormal{crude oil} \\
    \hdashline
     \textcolor{blue}{\textbf{Goal 2:} \textnormal{action}} \\
     \textcolor{teal}{\textbf{Reference:} \textnormal{cleanup efforts}} \\
    \textbf{Prediction (\texttt{Flan-T5-XL}): } \textnormal{accidents and safety}  \\
    \textbf{Prediction (\texttt{GPT-4o (zero-shot}): } \textnormal{criticized cleanup efforts, created a slick, ran hard aground, halted early, begin pumping, removing oil, placed a boom}  \\
    \textbf{Prediction (\texttt{GPT-4o (self-consistency}):} \textnormal{spread across, criticized cleanup efforts} \\
    \hline
    \hline
    \underline{\texttt{Source: DUC2001 (out-of-distribution)}} \\
    \textbf{Title: } (empty)  \\
    \textbf{Abstract: } \textnormal{the clinton administration will soon announce support for a north american development bank , which would fund projects in communities hit by job losses resulting from the north american free trade agreement . the so - called nadbank has been strongly supported by congressman esteban torres , who has insisted on some sort of lending institution to support adjustment throughout the continent . agreement by the administration is expected to bring mr torres and at least [digit] other hispanic congressmen into the pro - nafta fold . the administration believes it can garner [digit] - [digit] pro - nafta votes , out of the [digit] needed . }  \\
    \hdashline
    \textcolor{blue}{\textbf{Goal 1: } \textnormal{economic issue}}  \\
    \textcolor{teal}{\textbf{Reference:} \textnormal{job losses}} \\
    \textbf{Prediction (\texttt{Flan-T5-XL}): } \textnormal{jobs}  \\
    \textbf{Prediction (\texttt{GPT-4o (zero-shot}): } \textnormal{north american development bank, job losses, north american free trade agreement, lending institution, pro-nafta votes, anti-nafta public opinion}  \\
    \textbf{Prediction (\texttt{GPT-4o (self-consistency}):} \textnormal{north american development bank, clinton administration, job losses} \\
    \hdashline
     \textcolor{blue}{\textbf{Goal 2:}  \textnormal{political entity}} \\
     \textcolor{teal}{\textbf{Reference:} \textnormal{clinton administration}} \\
    \textbf{Prediction (\texttt{Flan-T5-XL}): }\textnormal{united states politics and government}  \\
    \textbf{Prediction (\texttt{GPT-4o (zero-shot}): } \textnormal{clinton administration, congressman esteban torres, hispanic congressmen, white house, president bill clinton}  \\
    \textbf{Prediction (\texttt{GPT-4o (self-consistency}):} \textnormal{clinton administration, north american development bank} \\
    \hline  
\end{tabular}
\caption{Examples of on-demand keyphrase generation instances and model outputs in the news domain.}
\label{qualitative-study-news}
\end{figure*}

\begin{figure*}[h!]
\small
\centering
\begin{tabular}{p{0.98\linewidth}}
    \hline
    \multicolumn{1}{c}{\texttt{Domain: Biomedical Text}} \\
    \hline
    \hline
    \underline{\texttt{Source: KPBiomed}} \\
    \textbf{Title:} \textnormal{contemporary trend of acute kidney injury incidence and incremental costs among us patients undergoing percutaneous coronary procedures .} \\
    \textbf{Abstract:} \textnormal{objectives to assess national trends of acute kidney injury ( aki ) incidence , incremental costs , risk factors , and readmissions among patients undergoing coronary angiography ( cag ) and / or percutaneous coronary intervention ( pci ) during [digit] - [digit] . background aki remains a serious complication for patients undergoing cag / pci . evidence is lacking in contemporary aki trends and its impact on hospital resource utilization . methods patients who underwent cag / pci procedures in [digit] hospitals were identified from premier healthcare database . aki was defined by icd - [digit] / [digit] diagnosis codes ( [digit] .} \\
    \hdashline
    \textcolor{blue}{\textbf{Goal 1:}  \textnormal{medical condition}} \\
    \textcolor{teal}{\textbf{Reference:} \textnormal{acute kidney injury, chronic kidney disease, nephropathy}} \\
    \textbf{Prediction (\texttt{Flan-T5-XL}):}  \textnormal{acute kidney injury}\\
    \textbf{Prediction (\texttt{GPT-4o (zero-shot}):} \textnormal{acute kidney injury, chronic kidney disease, anemia, diabetes} \\
    \textbf{Prediction (\texttt{GPT-4o (self-consistency}):}  \textnormal{acute kidney injury, chronic kidney disease, anemia}\\
    \hdashline
    \textcolor{blue}{\textbf{Goal 2:}  \textnormal{medical procedure}}\\
    \textcolor{teal}{\textbf{Reference:} \textnormal{percutaneous coronary intervention}} \\
    \textbf{Prediction (\texttt{Flan-T5-XL}):}  \textnormal{percutaneous coronary intervention}\\
    \textbf{Prediction (\texttt{GPT-4o (zero-shot}):} \textnormal{percutaneous coronary intervention, coronary angiography, coronary procedures, inpatient procedures, outpatient procedure} \\
    \textbf{Prediction (\texttt{GPT-4o (self-consistency}):}  \textnormal{percutaneous coronary intervention, coronary angiography}\\
    \hline
    \hline
    \underline{\texttt{Source: PubMed (out-of-distribution)}} \\
    \textbf{Title:} \textnormal{surviving sepsis campaign : international guidelines for management of severe sepsis and septic shock : [digit]} \\
     \textbf{Abstract:} \textnormal{objective to provide an update to the original surviving sepsis campaign clinical management guidelines , \u201c surviving sepsis campaign guidelines for management of severe sepsis and septic shock ,\u201d published in [digit] . introduction severe sepsis ( acute organ dysfunction secondary to infection ) and septic shock ( severe sepsis plus hypotension not reversed with fluid resuscitation ) are major healthcare problems , affecting millions of individuals around the world each year , killing one in four ( and often more ), and increasing in incidence [ [digit] \u2013 [digit] ]. similar to polytrauma , acute myocardial infarction , or stroke , the speed and appropriateness of therapy administered in the initial hours after severe sepsis develops are likely to influence outcome . } \\
    \textcolor{blue}{\textbf{Goal 1:}  \textnormal{medical condition}} \\
    \textcolor{teal}{\textbf{Reference:} \textnormal{sepsis, severe sepsis, septic shock, sepsis syndrome, infection}} \\
    \textbf{Prediction (\texttt{Flan-T5-XL}):}  \textnormal{sepsis}\\
    \textbf{Prediction (\texttt{GPT-4o (zero-shot}):} \textnormal{acute kidney injury, chronic kidney disease, anemia, diabetes} \\
    \textbf{Prediction (\texttt{GPT-4o (self-consistency}):}  \textnormal{severe sepsis, septic shock}\\
    \hdashline
    \textcolor{blue}{\textbf{Goal 2:}  \textnormal{healthcare initiative}}\\
    \textcolor{teal}{\textbf{Reference:} \textnormal{surviving sepsis campaign}} \\
    \textbf{Prediction (\texttt{Flan-T5-XL}):}  \textnormal{surviving sepsis campaign}\\
    \textbf{Prediction (\texttt{GPT-4o (zero-shot}):} \textnormal{surviving sepsis campaign, international guidelines, management of severe sepsis, septic shock, clinical management guidelines, evidence-based methodology} \\
    \textbf{Prediction (\texttt{GPT-4o (self-consistency}):}  \textnormal{surviving sepsis campaign}\\
    \hline
    \hline
    \underline{\texttt{Source: PubMed (out-of-distribution)}} \\
    \textbf{Title:} \textnormal{keratinocyte serum - free medium maintains long - term liver gene expression and function in cultured rat hepatocytes by preventing the loss of liver - enriched transcription factors} \\
    \textbf{Abstract:} \textnormal{freshly isolated hepatocytes rapidly lose their differentiated properties when placed in culture . therefore , production of a simple culture system for maintaining the phenotype of hepatocytes in culture would greatly facilitate their study . our aim was to identify conditions that could maintain the differentiated properties of hepatocytes for up to [digit] days of culture . adult rat hepatocytes were isolated and attached in williams \u2019 medium e containing [digit] \% serum . the medium was changed to either fresh williams \u2019 medium e or keratinocyte serum - free medium supplemented with dexamethasone , epidermal growth factor and pituitary gland extract .} \\
    \hdashline
    \textcolor{blue}{\textbf{Goal 1:}  \textnormal{biological extract} } \\
    \textcolor{teal}{\textbf{Reference:} \textnormal{pituitary gland extract}} \\
    \textbf{Prediction (\texttt{Flan-T5-XL}):}  \textnormal{pituitary gland extract}\\
    \textbf{Prediction (\texttt{GPT-4o (zero-shot}):} \textnormal{keratinocyte serum-free medium, Williams\u2019 medium E, dexamethasone, epidermal growth factor, pituitary gland extract} \\
    \textbf{Prediction (\texttt{GPT-4o (self-consistency}):}  \textnormal{keratinocyte serum-free medium, keratinocyte serum}\\
    \hdashline
    \textcolor{blue}{\textbf{Goal 2:}  \textnormal{molecular biology technique}} \\
    \textcolor{teal}{\textbf{Reference:} \textnormal{reverse transcription polymerase chain reaction}} \\
    \textbf{Prediction (\texttt{Flan-T5-XL}):}  \textnormal{cell culture}\\
    \textbf{Prediction (\texttt{GPT-4o (zero-shot}):} \textnormal{immunohistochemistry, western blotting, rt-pcr, immunofluorescence staining, collagenase perfusion technique} \\
    \textbf{Prediction (\texttt{GPT-4o (self-consistency}):}  \textnormal{western blotting, rt-pcr, immunohistochemistry}\\ 
    \hline  

    \hline  
    
\end{tabular}
\caption{Examples of on-demand keyphrase generation instances and model outputs in the biomedical domain.}
\label{qualitative-study-biomed}
\end{figure*}

%% file: main.bbl
\begin{thebibliography}{42}
\providecommand{\natexlab}[1]{#1}

\bibitem[{Augenstein et~al.(2017)Augenstein, Das, Riedel, Vikraman, and McCallum}]{augenstein-etal-2017-semeval}
Isabelle Augenstein, Mrinal Das, Sebastian Riedel, Lakshmi Vikraman, and Andrew McCallum. 2017.
\newblock \href {https://doi.org/10.18653/v1/S17-2091} {{S}em{E}val 2017 task 10: {S}cience{IE} - extracting keyphrases and relations from scientific publications}.
\newblock In \emph{Proceedings of the 11th International Workshop on Semantic Evaluation ({S}em{E}val-2017)}, pages 546--555, Vancouver, Canada. Association for Computational Linguistics.

\bibitem[{Berend(2011)}]{berend-2011-opinion}
G{\'a}bor Berend. 2011.
\newblock \href {https://aclanthology.org/I11-1130} {Opinion expression mining by exploiting keyphrase extraction}.
\newblock In \emph{Proceedings of 5th International Joint Conference on Natural Language Processing}, pages 1162--1170, Chiang Mai, Thailand. Asian Federation of Natural Language Processing.

\bibitem[{Boudin et~al.(2020)Boudin, Gallina, and Aizawa}]{boudin-etal-2020-keyphrase}
Florian Boudin, Ygor Gallina, and Akiko Aizawa. 2020.
\newblock \href {https://doi.org/10.18653/v1/2020.acl-main.105} {Keyphrase generation for scientific document retrieval}.
\newblock In \emph{Proceedings of the 58th Annual Meeting of the Association for Computational Linguistics}, pages 1118--1126, Online. Association for Computational Linguistics.

\bibitem[{Chen et~al.(2020)Chen, Chan, Li, and King}]{chen-etal-2020-exclusive}
Wang Chen, Hou~Pong Chan, Piji Li, and Irwin King. 2020.
\newblock \href {https://doi.org/10.18653/v1/2020.acl-main.103} {Exclusive hierarchical decoding for deep keyphrase generation}.
\newblock In \emph{Proceedings of the 58th Annual Meeting of the Association for Computational Linguistics}, pages 1095--1105, Online. Association for Computational Linguistics.

\bibitem[{Dou et~al.(2021)Dou, Liu, Hayashi, Jiang, and Neubig}]{dou-etal-2021-gsum}
Zi-Yi Dou, Pengfei Liu, Hiroaki Hayashi, Zhengbao Jiang, and Graham Neubig. 2021.
\newblock \href {https://doi.org/10.18653/v1/2021.naacl-main.384} {{GS}um: A general framework for guided neural abstractive summarization}.
\newblock In \emph{Proceedings of the 2021 Conference of the North American Chapter of the Association for Computational Linguistics: Human Language Technologies}, pages 4830--4842, Online. Association for Computational Linguistics.

\bibitem[{Feng et~al.(2024)Feng, Shi, Wang, Ding, Balachandran, and Tsvetkov}]{feng2024don}
Shangbin Feng, Weijia Shi, Yike Wang, Wenxuan Ding, Vidhisha Balachandran, and Yulia Tsvetkov. 2024.
\newblock \href {https://doi.org/10.18653/v1/2024.acl-long.786} {Don{'}t hallucinate, abstain: Identifying {LLM} knowledge gaps via multi-{LLM} collaboration}.
\newblock In \emph{Proceedings of the 62nd Annual Meeting of the Association for Computational Linguistics (Volume 1: Long Papers)}, pages 14664--14690, Bangkok, Thailand. Association for Computational Linguistics.

\bibitem[{Gallina et~al.(2019)Gallina, Boudin, and Daille}]{gallina-etal-2019-kptimes}
Ygor Gallina, Florian Boudin, and Beatrice Daille. 2019.
\newblock \href {https://doi.org/10.18653/v1/W19-8617} {{KPT}imes: A large-scale dataset for keyphrase generation on news documents}.
\newblock In \emph{Proceedings of the 12th International Conference on Natural Language Generation}, pages 130--135, Tokyo, Japan. Association for Computational Linguistics.

\bibitem[{Houbre et~al.(2022)Houbre, Boudin, and Daille}]{houbre-etal-2022-large}
Ma{\"e}l Houbre, Florian Boudin, and Beatrice Daille. 2022.
\newblock \href {https://doi.org/10.18653/v1/2022.louhi-1.6} {A large-scale dataset for biomedical keyphrase generation}.
\newblock In \emph{Proceedings of the 13th International Workshop on Health Text Mining and Information Analysis (LOUHI)}, pages 47--53, Abu Dhabi, United Arab Emirates (Hybrid). Association for Computational Linguistics.

\bibitem[{Hulth(2003)}]{hulth-2003-improved}
Anette Hulth. 2003.
\newblock \href {https://aclanthology.org/W03-1028} {Improved automatic keyword extraction given more linguistic knowledge}.
\newblock In \emph{Proceedings of the 2003 Conference on Empirical Methods in Natural Language Processing}, pages 216--223.

\bibitem[{Jiao et~al.(2023)Jiao, Zhong, Li, Zhao, Ouyang, Ji, and Han}]{jiao-etal-2023-instruct}
Yizhu Jiao, Ming Zhong, Sha Li, Ruining Zhao, Siru Ouyang, Heng Ji, and Jiawei Han. 2023.
\newblock \href {https://doi.org/10.18653/v1/2023.emnlp-main.620} {Instruct and extract: Instruction tuning for on-demand information extraction}.
\newblock In \emph{Proceedings of the 2023 Conference on Empirical Methods in Natural Language Processing}, pages 10030--10051, Singapore. Association for Computational Linguistics.

\bibitem[{Kim et~al.(2013)Kim, Kim, Cattle, Otmakhova, Park, and Shin}]{kim-etal-2013-applying}
Youngsam Kim, Munhyong Kim, Andrew Cattle, Julia Otmakhova, Suzi Park, and Hyopil Shin. 2013.
\newblock \href {https://aclanthology.org/I13-1108} {Applying graph-based keyword extraction to document retrieval}.
\newblock In \emph{Proceedings of the Sixth International Joint Conference on Natural Language Processing}, pages 864--868, Nagoya, Japan. Asian Federation of Natural Language Processing.

\bibitem[{Kulkarni et~al.(2022)Kulkarni, Mahata, Arora, and Bhowmik}]{kulkarni-etal-2022-learning}
Mayank Kulkarni, Debanjan Mahata, Ravneet Arora, and Rajarshi Bhowmik. 2022.
\newblock \href {https://doi.org/10.18653/v1/2022.findings-naacl.67} {Learning rich representation of keyphrases from text}.
\newblock In \emph{Findings of the Association for Computational Linguistics: NAACL 2022}, pages 891--906, Seattle, United States. Association for Computational Linguistics.

\bibitem[{Lewis et~al.(2020)Lewis, Liu, Goyal, Ghazvininejad, Mohamed, Levy, Stoyanov, and Zettlemoyer}]{lewis-etal-2020-bart}
Mike Lewis, Yinhan Liu, Naman Goyal, Marjan Ghazvininejad, Abdelrahman Mohamed, Omer Levy, Veselin Stoyanov, and Luke Zettlemoyer. 2020.
\newblock \href {https://doi.org/10.18653/v1/2020.acl-main.703} {{BART}: Denoising sequence-to-sequence pre-training for natural language generation, translation, and comprehension}.
\newblock In \emph{Proceedings of the 58th Annual Meeting of the Association for Computational Linguistics}, pages 7871--7880, Online. Association for Computational Linguistics.

\bibitem[{Li et~al.(2020)Li, Li, Mou, Jiang, Lyu, and King}]{DBLP:conf/nips/LiLMJLK20}
Jingjing Li, Zichao Li, Lili Mou, Xin Jiang, Michael~R. Lyu, and Irwin King. 2020.
\newblock \href {https://proceedings.neurips.cc/paper/2020/hash/7a677bb4477ae2dd371add568dd19e23-Abstract.html} {Unsupervised text generation by learning from search}.
\newblock In \emph{Advances in Neural Information Processing Systems 33: Annual Conference on Neural Information Processing Systems 2020, NeurIPS 2020, December 6-12, 2020, virtual}.

\bibitem[{Longpre et~al.(2023)Longpre, Hou, Vu, Webson, Chung, Tay, Zhou, Le, Zoph, Wei, and Roberts}]{pmlr-v202-longpre23a}
Shayne Longpre, Le~Hou, Tu~Vu, Albert Webson, Hyung~Won Chung, Yi~Tay, Denny Zhou, Quoc~V Le, Barret Zoph, Jason Wei, and Adam Roberts. 2023.
\newblock \href {https://proceedings.mlr.press/v202/longpre23a.html} {The flan collection: Designing data and methods for effective instruction tuning}.
\newblock In \emph{Proceedings of the 40th International Conference on Machine Learning}, volume 202 of \emph{Proceedings of Machine Learning Research}, pages 22631--22648. PMLR.

\bibitem[{Luan et~al.(2018)Luan, He, Ostendorf, and Hajishirzi}]{luan-etal-2018-multi}
Yi~Luan, Luheng He, Mari Ostendorf, and Hannaneh Hajishirzi. 2018.
\newblock \href {https://doi.org/10.18653/v1/D18-1360} {Multi-task identification of entities, relations, and coreference for scientific knowledge graph construction}.
\newblock In \emph{Proceedings of the 2018 Conference on Empirical Methods in Natural Language Processing}, pages 3219--3232, Brussels, Belgium. Association for Computational Linguistics.

\bibitem[{Meng et~al.(2017)Meng, Zhao, Han, He, Brusilovsky, and Chi}]{meng-etal-2017-deep}
Rui Meng, Sanqiang Zhao, Shuguang Han, Daqing He, Peter Brusilovsky, and Yu~Chi. 2017.
\newblock \href {https://doi.org/10.18653/v1/P17-1054} {Deep keyphrase generation}.
\newblock In \emph{Proceedings of the 55th Annual Meeting of the Association for Computational Linguistics (Volume 1: Long Papers)}, pages 582--592, Vancouver, Canada. Association for Computational Linguistics.

\bibitem[{OpenAI(2023)}]{OpenAI2023GPT4TR}
OpenAI. 2023.
\newblock Gpt-4 technical report.
\newblock \emph{ArXiv}, abs/2303.08774.

\bibitem[{Ouyang et~al.(2022)Ouyang, Wu, Jiang, Almeida, Wainwright, Mishkin, Zhang, Agarwal, Slama, Ray et~al.}]{ouyang2022training}
Long Ouyang, Jeffrey Wu, Xu~Jiang, Diogo Almeida, Carroll Wainwright, Pamela Mishkin, Chong Zhang, Sandhini Agarwal, Katarina Slama, Alex Ray, et~al. 2022.
\newblock \href {https://proceedings.neurips.cc/paper_files/paper/2022/file/b1efde53be364a73914f58805a001731-Paper-Conference.pdf} {Training language models to follow instructions with human feedback}.
\newblock \emph{Advances in neural information processing systems}, 35:27730--27744.

\bibitem[{Parekh et~al.(2024)Parekh, Mac, Yu, Dong, Shahriar, Liu, Yang, Huang, Wang, Peng, and Chang}]{parekh2024event}
Tanmay Parekh, Anh Mac, Jiarui Yu, Yuxuan Dong, Syed Shahriar, Bonnie Liu, Eric Yang, Kuan-Hao Huang, Wei Wang, Nanyun Peng, and Kai-Wei Chang. 2024.
\newblock \href {https://arxiv.org/abs/2404.01679} {Event detection from social media for epidemic prediction}.
\newblock \emph{Preprint}, arXiv:2404.01679.

\bibitem[{Park and Caragea(2023)}]{park-caragea-2023-multi}
Seo Park and Cornelia Caragea. 2023.
\newblock \href {https://doi.org/10.18653/v1/2023.emnlp-main.805} {Multi-task knowledge distillation with embedding constraints for scholarly keyphrase boundary classification}.
\newblock In \emph{Proceedings of the 2023 Conference on Empirical Methods in Natural Language Processing}, pages 13026--13042, Singapore. Association for Computational Linguistics.

\bibitem[{Park and Caragea(2020)}]{park-caragea-2020-scientific}
Seoyeon Park and Cornelia Caragea. 2020.
\newblock \href {https://doi.org/10.18653/v1/2020.coling-main.472} {Scientific keyphrase identification and classification by pre-trained language models intermediate task transfer learning}.
\newblock In \emph{Proceedings of the 28th International Conference on Computational Linguistics}, pages 5409--5419, Barcelona, Spain (Online). International Committee on Computational Linguistics.

\bibitem[{QasemiZadeh and Schumann(2016)}]{qasemizadeh-schumann-2016-acl}
Behrang QasemiZadeh and Anne-Kathrin Schumann. 2016.
\newblock \href {https://aclanthology.org/L16-1294} {The {ACL} {RD}-{TEC} 2.0: A language resource for evaluating term extraction and entity recognition methods}.
\newblock In \emph{Proceedings of the Tenth International Conference on Language Resources and Evaluation ({LREC}'16)}, pages 1862--1868, Portoro{\v{z}}, Slovenia. European Language Resources Association (ELRA).

\bibitem[{Schutz(2008)}]{Schutz2008KeyphraseEF}
Alexander Schutz. 2008.
\newblock \href {https://api.semanticscholar.org/CorpusID:8314070} {Keyphrase extraction from single documents in the open domain exploiting linguistic and statistical methods}.

\bibitem[{Song et~al.(2024)Song, Feng, and Jing}]{DBLP:journals/corr/abs-2405-16571}
Mingyang Song, Yi~Feng, and Liping Jing. 2024.
\newblock \href {https://doi.org/10.48550/ARXIV.2405.16571} {A preliminary empirical study on prompt-based unsupervised keyphrase extraction}.
\newblock \emph{CoRR}, abs/2405.16571.

\bibitem[{Song et~al.(2023)Song, Jiang, Shi, Yao, Lu, Feng, Liu, and Jing}]{song2023chatgpt}
Mingyang Song, Haiyun Jiang, Shuming Shi, Songfang Yao, Shilong Lu, Yi~Feng, Huafeng Liu, and Liping Jing. 2023.
\newblock Is chatgpt a good keyphrase generator? a preliminary study.
\newblock \emph{arXiv preprint arXiv:2303.13001}.

\bibitem[{Tang et~al.(2017)Tang, Huang, Liu, Tung, Wang, Yang, and Zhang}]{Tang2017QALinkET}
Yixuan Tang, Weilong Huang, Qi~Liu, Anthony K.~H. Tung, Xiaoli Wang, Jisong Yang, and Beibei Zhang. 2017.
\newblock \href {https://leuchine.github.io/papers/cikm17.pdf} {Qalink: Enriching text documents with relevant q\&a site contents}.
\newblock \emph{Proceedings of the 2017 ACM on Conference on Information and Knowledge Management}.

\bibitem[{Wan and Xiao(2008)}]{wan2008single}
Xiaojun Wan and Jianguo Xiao. 2008.
\newblock Single document keyphrase extraction using neighborhood knowledge.
\newblock In \emph{AAAI}, volume~8, pages 855--860.

\bibitem[{Wang et~al.(2023)Wang, Wei, Schuurmans, Le, Chi, Narang, Chowdhery, and Zhou}]{wang2023selfconsistency}
Xuezhi Wang, Jason Wei, Dale Schuurmans, Quoc~V Le, Ed~H. Chi, Sharan Narang, Aakanksha Chowdhery, and Denny Zhou. 2023.
\newblock \href {https://openreview.net/forum?id=1PL1NIMMrw} {Self-consistency improves chain of thought reasoning in language models}.
\newblock In \emph{The Eleventh International Conference on Learning Representations}.

\bibitem[{Wang et~al.(2019)Wang, Li, Chan, King, Lyu, and Shi}]{wang-etal-2019-topic-aware}
Yue Wang, Jing Li, Hou~Pong Chan, Irwin King, Michael~R. Lyu, and Shuming Shi. 2019.
\newblock \href {https://doi.org/10.18653/v1/P19-1240} {Topic-aware neural keyphrase generation for social media language}.
\newblock In \emph{Proceedings of the 57th Annual Meeting of the Association for Computational Linguistics}, pages 2516--2526, Florence, Italy. Association for Computational Linguistics.

\bibitem[{Wang et~al.(2016)Wang, Jin, Zhu, and Goutte}]{wang-etal-2016-extracting}
Yunli Wang, Yong Jin, Xiaodan Zhu, and Cyril Goutte. 2016.
\newblock \href {https://aclanthology.org/C16-1089} {Extracting discriminative keyphrases with learned semantic hierarchies}.
\newblock In \emph{Proceedings of {COLING} 2016, the 26th International Conference on Computational Linguistics: Technical Papers}, pages 932--942, Osaka, Japan. The COLING 2016 Organizing Committee.

\bibitem[{Witten et~al.(1999)Witten, Paynter, Frank, Gutwin, and Nevill-Manning}]{witten1999kea}
Ian~H Witten, Gordon~W Paynter, Eibe Frank, Carl Gutwin, and Craig~G Nevill-Manning. 1999.
\newblock Kea: Practical automatic keyphrase extraction.
\newblock In \emph{Proceedings of the fourth ACM conference on Digital libraries}, pages 254--255.

\bibitem[{Wu et~al.(2023{\natexlab{a}})Wu, Ahmad, and Chang}]{wu-etal-2023-rethinking-model}
Di~Wu, Wasi Ahmad, and Kai-Wei Chang. 2023{\natexlab{a}}.
\newblock \href {https://doi.org/10.18653/v1/2023.emnlp-main.410} {Rethinking model selection and decoding for keyphrase generation with pre-trained sequence-to-sequence models}.
\newblock In \emph{Proceedings of the 2023 Conference on Empirical Methods in Natural Language Processing}, pages 6642--6658, Singapore. Association for Computational Linguistics.

\bibitem[{Wu et~al.(2022)Wu, Ahmad, Dev, and Chang}]{wu-etal-2022-representation}
Di~Wu, Wasi Ahmad, Sunipa Dev, and Kai-Wei Chang. 2022.
\newblock \href {https://doi.org/10.18653/v1/2022.findings-emnlp.49} {Representation learning for resource-constrained keyphrase generation}.
\newblock In \emph{Findings of the Association for Computational Linguistics: EMNLP 2022}, pages 700--716, Abu Dhabi, United Arab Emirates. Association for Computational Linguistics.

\bibitem[{Wu et~al.(2023{\natexlab{b}})Wu, Yin, and Chang}]{wu2023kpeval}
Di~Wu, Da~Yin, and Kai-Wei Chang. 2023{\natexlab{b}}.
\newblock \href {https://arxiv.org/abs/2303.15422} {Kpeval: Towards fine-grained semantic-based evaluation of keyphrase extraction and generation systems}.

\bibitem[{Yao et~al.(2019)Yao, Peng, Weischedel, Knight, Zhao, and Yan}]{yao2019plan}
Lili Yao, Nanyun Peng, Ralph Weischedel, Kevin Knight, Dongyan Zhao, and Rui Yan. 2019.
\newblock \href {https://ojs.aaai.org/index.php/AAAI/article/view/4726/4604} {Plan-and-write: Towards better automatic storytelling}.
\newblock In \emph{Proceedings of the AAAI Conference on Artificial Intelligence}, volume~33, pages 7378--7385.

\bibitem[{Zhang et~al.(2023)Zhang, Liu, Yang, Fang, Chen, Radev, Zhu, Zeng, and Zhang}]{zhang-etal-2023-macsum}
Yusen Zhang, Yang Liu, Ziyi Yang, Yuwei Fang, Yulong Chen, Dragomir Radev, Chenguang Zhu, Michael Zeng, and Rui Zhang. 2023.
\newblock \href {https://doi.org/10.1162/tacl_a_00575} {{MACS}um: Controllable summarization with mixed attributes}.
\newblock \emph{Transactions of the Association for Computational Linguistics}, 11:787--803.

\bibitem[{Zhang et~al.(2022{\natexlab{a}})Zhang, Jiang, Yang, Li, and Wang}]{DBLP:conf/sigir/ZhangJY0W22}
Yuxiang Zhang, Tao Jiang, Tianyu Yang, Xiaoli Li, and Suge Wang. 2022{\natexlab{a}}.
\newblock \href {https://doi.org/10.1145/3477495.3531990} {{HTKG:} deep keyphrase generation with neural hierarchical topic guidance}.
\newblock In \emph{{SIGIR} '22: The 45th International {ACM} {SIGIR} Conference on Research and Development in Information Retrieval, Madrid, Spain, July 11 - 15, 2022}, pages 1044--1054. {ACM}.

\bibitem[{Zhang et~al.(2022{\natexlab{b}})Zhang, Yang, Jiang, Li, and Wang}]{DBLP:conf/pkdd/ZhangYJLW22}
Yuxiang Zhang, Tianyu Yang, Tao Jiang, Xiaoli Li, and Suge Wang. 2022{\natexlab{b}}.
\newblock \href {https://doi.org/10.1007/978-3-031-26390-3\_30} {Hyperbolic deep keyphrase generation}.
\newblock In \emph{Machine Learning and Knowledge Discovery in Databases - European Conference, {ECML} {PKDD} 2022, Grenoble, France, September 19-23, 2022, Proceedings, Part {II}}, volume 13714 of \emph{Lecture Notes in Computer Science}, pages 521--536. Springer.

\bibitem[{Zhao et~al.(2022)Zhao, Yin, Yang, and Yao}]{zhao-etal-2022-keyphrase}
Guangzhen Zhao, Guoshun Yin, Peng Yang, and Yu~Yao. 2022.
\newblock \href {https://doi.org/10.18653/v1/2022.emnlp-main.529} {Keyphrase generation via soft and hard semantic corrections}.
\newblock In \emph{Proceedings of the 2022 Conference on Empirical Methods in Natural Language Processing}, pages 7757--7768, Abu Dhabi, United Arab Emirates. Association for Computational Linguistics.

\bibitem[{Zhao et~al.(2021)Zhao, Song, Feng, Zhuang, Li, Wang, and Liu}]{deepkpcompletion}
Yu~Zhao, Jia Song, Huali Feng, Fuzhen Zhuang, Qing Li, Xiaojie Wang, and Ji~Liu. 2021.
\newblock \href {https://arxiv.org/abs/2111.01910} {Deep keyphrase completion}.
\newblock \emph{CoRR}, abs/2111.01910.

\bibitem[{Zhong et~al.(2021)Zhong, Yin, Yu, Zaidi, Mutuma, Jha, Awadallah, Celikyilmaz, Liu, Qiu, and Radev}]{zhong-etal-2021-qmsum}
Ming Zhong, Da~Yin, Tao Yu, Ahmad Zaidi, Mutethia Mutuma, Rahul Jha, Ahmed~Hassan Awadallah, Asli Celikyilmaz, Yang Liu, Xipeng Qiu, and Dragomir Radev. 2021.
\newblock \href {https://doi.org/10.18653/v1/2021.naacl-main.472} {{QMS}um: A new benchmark for query-based multi-domain meeting summarization}.
\newblock In \emph{Proceedings of the 2021 Conference of the North American Chapter of the Association for Computational Linguistics: Human Language Technologies}, pages 5905--5921, Online. Association for Computational Linguistics.

\end{thebibliography}
